\documentclass[pdflatex,sn-mathphys-num]{sn-jnl}% Math and Physical Sciences Numbered Reference Style
\usepackage{graphicx}%
\usepackage{subfigure}
\usepackage{multirow}%
\usepackage{multicol}%
\usepackage{amsmath,amssymb,amsfonts}%
\usepackage{amsthm}%
\usepackage{mathrsfs}%
\usepackage[title]{appendix}%
\usepackage{xcolor}%
\usepackage{textcomp}%
\usepackage{manyfoot}%
\usepackage{booktabs}%
\usepackage{algorithm}%
\usepackage{algorithmicx}%
\usepackage{algpseudocode}%
\usepackage{listings}%

\theoremstyle{thmstyleone}%
\theoremstyle{thmstyletwo}%

\theoremstyle{thmstylethree}%

\begin{document}

\title[Article Title]{Downsized and Compromised?: Assessing the Faithfulness of Model Compression}


\author*[1]{\fnm{Moumita} \sur{Kamal}}\email{mkamal42@tntech.edu}

\author[2]{\fnm{Douglas A.} \sur{Talbert}}\email{dtalbert@tntech.edu}
\equalcont{These authors contributed equally to this work.}

\affil*[1]{\orgdiv{Computer Science}, \orgname{Tennessee Tech University}, \orgaddress{ \city{Cookeville}, \state{TN}, \country{USA}}}

\affil[2]{\orgdiv{Computer Science}, \orgname{Tennessee Tech University}, \orgaddress{ \city{Cookeville}, \state{TN}, \country{USA}}}


\abstract{In real-world applications, computational constraints often require transforming large models into smaller, more efficient versions through model compression. While these techniques aim to reduce size and computational cost without sacrificing performance, their evaluations have traditionally focused on the trade-off between size and accuracy, overlooking the aspect of model faithfulness. This limited view is insufficient for high-stakes domains like healthcare, finance, and criminal justice, where compressed models must remain faithful to the behavior of their original counterparts. This paper presents a novel approach to evaluating faithfulness in compressed models, moving beyond standard metrics. We introduce and demonstrate a set of faithfulness metrics that capture how model behavior changes post-compression. Our contributions include introducing techniques to assess predictive consistency between the original and compressed models using model agreement, and applying chi-squared tests to detect statistically significant changes in predictive patterns across both the overall dataset and demographic subgroups, thereby exposing shifts that aggregate fairness metrics may obscure. We demonstrate our approaches by applying quantization and pruning to artificial neural networks (ANNs) trained on three diverse and socially meaningful datasets. Our findings show that high accuracy does not guarantee faithfulness, and our statistical tests detect subtle yet significant shifts that are missed by standard metrics, such as Accuracy and Equalized Odds. The proposed metrics provide a practical and more direct method for ensuring that efficiency gains through compression do not compromise the fairness or faithfulness essential for trustworthy AI.}




\maketitle

\section{Introduction}\label{intro}

As machine learning (ML) models continue to grow in size and complexity, their deployment in real-world applications is becoming increasingly challenging due to the limitations of edge devices, such as smartphones, embedded systems, and Internet of Things (IoT) hardware \cite{murshed2021machine}. These systems often have strict constraints on processing power, memory, battery life, and storage. Running large-scale deep learning models, particularly Artificial Neural Networks (ANNs), in such environments can be impractical. To address these challenges, model compression has emerged as an essential area of research. It focuses on reducing model size, improving inference latency, and enabling real-time deployment without compromising accuracy significantly \cite{buciluǎ2006model}.

While model compression has proven effective in enhancing resource efficiency, there is growing concern regarding the trustworthiness and ethical implications of deploying compressed models, particularly in sensitive areas such as healthcare, criminal justice, finance, and public policy \cite{hooker2019compressed}. Trustworthiness in machine learning encompasses issues like fairness, robustness, transparency, and consistency.  A key aspect of this is model faithfulness, which we define as the degree to which a compressed model's predictions and underlying behavior align with those of the original, uncompressed model. Recent studies have shown that model compression can exacerbate existing biases or introduce new forms of unfairness \cite{ramesh2023comparative, gonccalves2023understanding, stoychev2022effect}. In other words, while compressed models may perform well overall, they can behave more unfairly toward certain demographic groups or make different decisions compared to the original models. Despite the increasing use of compressed models in sensitive applications, relatively little work has been done to precisely evaluate the true alignment between compressed models and the original, uncompressed models, including measuring their impact on fairness. Thus, we still do not fully understand how these methods affect fairness or the faithfulness of predictions. This leaves a crucial gap in the research, particularly for applications where both performance and trust are essential.

This study aims to bridge this gap by introducing metrics to enable measuring both the efficiency and the faithfulness of compressed artificial neural network (ANN) models. We demonstrate these metrics using multiple datasets and multiple model compression techniques. Specifically, this paper makes the following contributions:
\begin{enumerate}
    \item We introduce novel approaches to quantify instance-level and subgroup-level model agreement.
    \item We illustrate a novel method that uses statistical tests to identify when there are significant differences between how models perform at both an instance-level and a subgroup-level.
    \item We demonstrate that the outcomes of our proposed metrics can be reliably predicted using validation sets, making them practical diagnostic tools.
\end{enumerate}

Through this work, we provide a more direct approach for evaluating compressed models, enabling a more faithful and trustworthy deployment of AI in resource-constrained environments.

The rest of the paper is organized as follows: Section \ref{sec:background} provides background on model compression techniques and introduces the concepts of trustworthy model compression and model bias. Section \ref{sec:related work} reviews related literature that has explored the effects of compression on model fairness. In Section \ref{sec:methodology}, we describe our experimental methodology, detailing the datasets, model architecture, and compression strategies employed. Section \ref{sec:accuracy} presents the results using traditional metrics of model size and accuracy, demonstrating that these metrics can be reliably predicted using a validation set. In Section \ref{sec:agreement}, we introduce our novel model agreement metric and use chi-squared tests to evaluate statistically significant shifts in model predictions following compression. Section \ref{sec:bias} extends this statistical analysis to model bias, evaluating how compression-induced changes affect fairness across demographic subgroups. Section \ref{sec:limitations} discusses the limitations of our approach, and Section \ref{sec:conclusion} concludes with a summary of our key findings and outlines directions for future research.

\section{Background}\label{sec:background}
\subsection{Model compression}
The primary notion behind model compression is to develop a technique that allows one to use a smaller and faster model to approximate the same function learned by a slower and bigger model \cite{buciluǎ2006model}. For example, the function learned by the larger, more accurate model can be used to label a significant amount of pseudo data, and by training a smaller, faster model on this data, we can minimize overfitting \cite{buciluǎ2006model}. This results in a compressed model that approximates the function of the larger model well while being faster and more efficient.

Model compression aims to reduce the computational cost of large neural networks or other ensemble models while minimizing the loss of accuracy. Pruning \cite{voita2019analyzing,prasanna2020bert}, quantization \cite{han2015deep,cheong2019transformers}, and knowledge distillation \cite{polino2018model,tan2018distill} techniques are some of the most commonly used model compression techniques \cite{gupta2022compression}. Pruning involves removing the less important model components (e.g., connections or neurons) from the model, whereas quantization reduces the precision of the model's weights. Knowledge distillation aims to transfer the knowledge learned by a complex model to a simpler model \cite{jiao2019tinybert}. The use of these techniques is becoming increasingly popular in the field of deep learning. Implementation of model compression allows us to reduce the model size significantly, making it more affordable and faster to execute \cite{deng2020model}. Additionally, this can lead to energy-efficient models that can be deployed on resource-limited devices, making them better suited for many real-world applications.

\subsection{Types of Model Compression}
\subsubsection{Pruning}
Pruning is a powerful technique used in machine learning to optimize neural network models \cite{dai2019grow}. When a neural network is trained, it often contains many relatively unimportant or redundant connections and neurons that do not significantly contribute to the model's performance. By selectively removing these connections and/or neurons, the overall size and complexity of the model can be reduced, resulting in a more consolidated and computationally efficient model \cite{he2014reshaping}. Pruning is popularly used as an essential tool for making neural networks more practical and efficient for a wide range of applications. Researchers commonly use two primary techniques for pruning: 

\textit{(1) Weight Pruning }- where we can remove connections in neurons by setting specific weights to zero \cite{han2015learning,guo2019reweighted}. 

\textit{(2) Node/Neuron Pruning} - where we can remove entire neurons from the neural network depending on their contributions toward the model's output \cite{murray2015auto} and activation patterns (i.e. how often they activate) \cite{pan2016dropneuron}. 

Pruning is often used as a means of creating sparsity in the weights and activations of a neural network \cite{dai2019grow}. The process entails training a neural network until convergence and ensuring a well-performing model. Next, the less important connections/neurons are identified (either iteratively or all at once) and removed. After pruning, the remaining parameters are often fine-tuned to improve performance \cite{gupta2022compression}.

\subsubsection{Quantization}
Small devices like microcontrollers and various wearable or edge devices have less memory capacity compared to a traditional computer. Quantization is a common technique to convert a large machine learning model into a smaller one so that it can be deployed on edge devices \cite{he2016effective}. Quantization reduces the number of bits used to represent each weight in a machine learning model by reducing its precision \cite{shen2020q}. In deep learning, the standard numerical format used for research and deployment is usually the 32-bit floating point (FP32). However, quantization experiments have shown that weights and activations can be defined using 8-bit integers (INT8) without incurring a significant loss in accuracy \cite{he2016effective}. By reducing the number of bits in each weight, the memory required to store the model is also reduced. 
This can be particularly useful when deploying machine learning models on resource-limited edge devices (e.g. a microcontroller with just a few megabytes of memory or a smartwatch). Moreover, quantization allows for a much faster inference time \cite{zhou2017balanced}. 

In this paper, we used \textit{quantization-aware training} - where quantization occurs during training as opposed to \textit{post-training quantization} - which involves quantizing a pre-trained model.

\subsubsection{Distillation}
Knowledge distillation is an advanced technique used to compress a larger, pre-trained machine learning model into a more compact form during the training process \cite{hinton2015distilling}. This concept was first introduced by Caruana et al. \cite{buciluǎ2006model} in 2006 and later generalized by Hinton et al. in 2015, making it an essential tool in the field of machine learning.

In distillation, the teacher model's knowledge is transferred to the student model by minimizing a loss function that uses the teacher model's predicted class probabilities \cite{hinton2015distilling}. In other words, it uses the output of a softmax function on the teacher model's logits. However, often, this distribution has only one highly probable class, providing little additional information beyond the dataset labels \cite{hinton2015distilling}.

\subsubsection{Low-rank approximation (LoRA)}
In addition to the methods discussed above, researchers talk about Low-rank Approximation (LoRA) \cite{yu2017compressing}, also known as tensor decomposition. This technique compresses a neural network model by reducing the rank of its weight matrices. It is based on the observation that the weight matrices of neural networks are often low rank, meaning they can be approximated by a product of two smaller matrices \cite{shu2017compressing,ye2018learning}. It can be particularly effective for compressing a model's large, fully connected layers. It's also known to be used in conjunction with other compression techniques, such as quantization (QLoRA), to enable faster fine-tuning \cite{sau2016deep}.  

Other techniques for model compression include Conditional Computation \cite{bengio2015conditional, bolukbasi2017adaptive}, Parameter Sharing \cite{tay2019lightweight, lan2019albert}, and Sub-Quadratic Complexity Transformers \cite{kitaev2020reformer}. However, in this paper, we have focused our experiments specifically on weight pruning and quantization for model compression.

\subsection{Trustworthy Model Compression} \label{sec:trustAcc}

Two metrics have been used almost exclusively for assessing the quality of model compression, \textit{model size} and \textit{model accuracy} \cite{cheng2017survey}. However, while \textit{accuracy} is important, it can hide differences in model performance. For example, two models could have the same accuracy with one having a high false positive rate and a low false negative rate and the other having a low false positive rate and a high false negative rate. Thus, while having the same accuracy, such models do not behave the same way. Thus, if one is seeking to measure the \textit{faithfulness} of a compressed model to the original uncompressed model, one should look beyond just accuracy and assess additional metrics, including \textit{model agreement} and \text{change in model bias}.

\subsubsection{Model Agreement}
\textit{Model agreement} can be analyzed directly by identifying and characterizing instances on which the uncompressed model and its compressed counterpart agree compared to those on which the two models disagree. Additionally, one can measure the statistical significance of any changes in the distribution of the predicted classes using a \textit{chi-squared test}.

The chi-squared test is a statistical method used to evaluate the association between two categorical variables \cite{ugoni1995chi}, with p-values indicating the significance of these associations. Chi-square statistics can also assess the performance of classification models by analyzing confusion matrices or comparing predicted and actual class distributions \cite{franke2012chi}.

To assess the faithfulness of model compression, a chi-squared test can be used to see if there is a statistically significant association between the model being compressed or not with the distribution of predicted classes. If the computed p-value indicates statistical significance, then there is statistical evidence that model compression resulted in a change in the distribution of predicted classes beyond that which would be expected due to randomness. This would support drawing the conclusion that the compressed model is not a faithful (trusted) compression. 

More details regarding the measurement of model agreement and the use of the chi-squared test are included in Section \ref{sec:agreement}.

\subsubsection{Change in Model Bias}
Another metric that can assess the faithfulness of a compressed model is a comparison of the uncompressed model's bias to that of the compressed model. While there is some evidence that smaller models can reduce bias in some cases, that bias reduction was accompanied by a decrease in accuracy \cite{talbert2024assessing}. This change could be measured by comparing overall model biases using an accepted bias metric such as \textit{equalized odds} \cite{hardt2016equality}.

Like accuracy, however, such metrics could hide changes in the details by shifting the bias around among groups. Thus, as with model agreement, faithfulness regarding bias is better captured by measuring how similarly the uncompressed and compressed models treat meaningful subsets of data more directly. 

Additional background on measuring model bias is found in Section \ref{sec:biasBG}, and more details regarding assessing faithfulness through change of model bias is included in Section \ref{sec:bias}.

\subsubsection{Other Faithfulness Metrics}

While not addressed in this paper, other metrics could also be considered to assess the faithfulness of a compress model. These include measuring the \textit{change in explanations} between the uncompressed and compressed model as well as measuring the change in \textit{classification uncertainty} between the uncompressed and compressed models.

\subsection{Model Bias} \label{sec:biasBG}
Bias in machine learning systems is a widely researched topic for which various concepts of fairness have been explored \cite{mehrabi2021survey,verma2018fairness}. Incidents of machine bias like \cite{angwin2022machine} and \cite{noble2018algorithms} adversely impacting certain groups have further added to the importance of this research. When evaluating a model's bias, one commonly used approach is to analyze its performance across different demographic groups (e.g., age, sex, race/ethnicity, location, etc.) \cite{dealcala2023measuring}. It provides insights into whether the model's predictions are consistent across the different groups or if errors negatively impact certain groups. 

There are several metrics to calculate the fairness of a model \cite{mehrabi2021survey,makhlouf2021applicability}. For example, one commonly used metric is \textit{treatment equality}, which evaluates whether the ratio of false negatives and false positives is the same for all groups, regardless of their demographic characteristics \cite{berk2021fairness}. Another metric is \textit{fairness through unawareness}, which deems an algorithm fair as long as no protected features are used in the decision-making process \cite{grgic2016case}. \textit{Demographic parity} is a metric that examines whether the model's predictions align with the proportion of each group in the overall population \cite{dwork2012fairness}. By using these metrics, one can identify biases in a model and make changes to improve its overall fairness and performance.

In this paper, we have used \textit{equalized odds} as our fairness metric \cite{phillips2023group}. This metric considers a model to be fair if the subgroups have equal true positive rates (TPR) and equal false positive rates (FPR). We can also define this metric using sensitivity (TPR) and specificity (1-FPR). We used the bias function introduced in \cite{phillips2023group} defined as follows: 

\begin{align}
\begin{split}
    &\text{bias}(M,G = \{g_1,\ldots,g_k\}) \\
    &= \sum_{i = 1}^{k}\sum_{j=i+1}^{k}|\text{Specificity}(M, g_j) - \text{Specificity}(M, g_i)| \\
    &+ \sum_{i = 1}^{k}\sum_{j=i+1}^{k}|\text{Sensitivity}(M, g_i) - \text{Sensitivity}(M, g_j)|
\end{split}
\label{eq:1}
\end{align}

where, $M$ is the machine learning model and $g_1, g_2, ..., g_k$ are subsets of demographic groups. The smaller this value is, the fairer the model. An algorithm is considered to be completely impartial relative to the considered demographic groups if this equation evaluates to 0.

\section{Related Work}\label{sec:related work}
Though underexplored, some research has been done on the effects of model compression on fairness/bias. Hooker et al. \cite{hooker2019compressed, hooker2021moving} were among the first to systematically show that compressed models, despite maintaining overall accuracy, can exhibit ``selective forgetting,'' disproportionately degrading performance for underrepresented classes or specific types of inputs. In their paper, Xu et al. discuss how distillation and pruning affect toxicity and bias in generative language models \cite{xu2022can}. According to the paper's findings, Xu et al. conclude that knowledge distillation produces less toxic and possibly less biased models \cite{xu2022can}. Inspired by the knowledge distillation literature, Joseph et al. propose a novel loss function for model compression \cite{joseph2020going}. While their study heavily focuses on prediction accuracy and the loss function, they do discuss the effects of their approach on model bias. According to the survey, their approach was able to preserve the model's fairness in most cases \cite{joseph2020going}. Stoychev and Gunes used model compression on a facial expression recognition algorithm \cite{stoychev2022effect}. They explore the effects of compression on accuracy and fairness. However, they measure fairness as the difference in accuracy in two bias groups \cite{stoychev2022effect}. Iofinova et al. \cite{iofinova2023bias} provided an in-depth analysis of bias in pruned vision models, noting increased uncertainty and correlations at higher sparsity, which they linked to increased bias. However, they use the Softmax function for their uncertainty quantification (UQ), which is often regarded as an unsatisfactory measure for UQ \cite{guo2017calibration}. Ramesh et al. \cite{ramesh2023comparative} conducted a comparative study on the impact of model compression in language model fairness, and they discovered that distilled models often displayed more bias in both intrinsic and extrinsic fairness measures compared to their original versions or even to pruned or quantized models.

\section{Experimental Methodology}\label{sec:methodology}
\subsection{The Datasets}
For our experiments, we used three datasets: the COMPAS dataset, the Employment dataset, and the Trauma dataset. Each dataset comes from a different domain and reflects real-world challenges and biases. We discuss the details of these datasets and the bias groups represented within them in depth below.
\subsubsection{COMPAS}
We obtained the data for our first dataset from the COMPAS Recidivism Racial Bias dataset analyzed by ProPublica \cite{angwin2022machine}. The Correctional Offender Management Profiling for Alternative Sanctions (COMPAS) is a criminal risk assessment tool widely used by parole officers and judges. This algorithm predicts and scores the likelihood of a criminal's recidivism or re-offense \cite{dressel2018accuracy}. Researchers have performed a two-year follow-up study on the criminal defendants to confirm the legitimacy of the algorithm and found that COMPAS is biased against African American defendants and in favor of Caucasian defendants \cite{angwin2022machine}. We chose this dataset specifically because its well-documented, real-world biases provide a critical testbed for our research. It allows us to explore the fairness measures of our model across different groups (e.g. race, age, sex) and, more importantly, to analyze what impact model compression has on the fairness/bias of a model.

Our dataset consisted of over 18,000 instances of criminal defendant information used by the COMPAS algorithm, along with the decision made by the algorithm and the outcome recorded after two years of the decision. The data was described using 34 features, including the defendant's demographic information, degree of charges, prior history and the COMPAS score. Additionally, each sample was labeled with the study outcome defined as a \textit{recid} value of 0 or 1. 

\begin{table}[ht]
\caption{Bias Subgroups in the COMPAS Dataset}
\centering
\begin{tabular}{lr} \hline
\textbf{Bias Group} & \textbf{Size} \\ \hline 
Race = African American & 10,074 \\ 
Race = Caucasian & 6,438 \\ [.35em]
Sex = Male & 13,465 \\ 
Sex = Female & 3,047 \\[.35em]
Age \(<\) 25 & 3,862 \\ 
Age = 25-45  & 9,423 \\ 
Age \(>\) 45 & 3,227 \\ \hline
\end{tabular}
\label{tbl_1}
\end{table}

We analyzed groups in our dataset to determine bias based on three demographic variables: sex, race, and age. The instances in our dataset were divided into six kinds according to race. They were Caucasian, African American, Asian, Hispanic, Native American and Others. Upon analysis, we discovered that Asians, Native Americans, Hispanics, and Others had relatively small instances compared to Caucasians and African Americans. Additionally, as some researchers have suggested that the COMPAS algorithm is biased towards Caucasians and against African American people, we decided to focus our research only on these two groups for this paper. Thus, we dropped the samples with the other four races (Asian, Native American, Hispanic, and Others) as attributes, resulting in a dataset with 16,512 instances.  Table \ref{tbl_1} lists the seven subgroups and their sizes. 

\subsubsection{Trauma Data}
Our second dataset was extracted from the trauma registry of a Level I Trauma Center, spanning from 1991 to 2016. This comprehensive registry includes all trauma patients aged 16 and older who were treated at the facility. The dataset is thorough, comprising 32 distinct features that capture various aspects of each case, including patient demographics, vital physiological parameters, specific anatomical criteria, and the mechanisms of injury sustained. In addition to these features, each data point is labeled with the target class: ``severely injured," which is defined by an Injury Severity Score (ISS) exceeding 15. This classification highlights the extent of trauma sustained by the patients in our study. We selected this dataset to evaluate our faithfulness metrics in a critical medical decision-making context where model predictions could directly influence patient care pathways. We focused our analysis on a cohort of 50,644 individuals, all of whom had complete initial physiological values recorded during their emergency department visit. To explore potential biases within our dataset, we examined differences across two key demographic variables: age and sex. Analyzing these traditional axes of concern is essential for ensuring that predictive models used in trauma care are not only accurate but also fair, as any systematic errors introduced by model compression could have profound consequences on patient triage and treatment.

\begin{table}[ht]
\caption{Bias Subgroups in the Trauma Dataset}
\centering
\begin{tabular}{lr} \toprule
\textbf{Bias Group} & \textbf{Size} \\ \midrule
Sex = M & 34,577 \\
Sex = F & 16,067 \\ [.35em]
Age \(<\) 65 & 42,912 \\ 
Age \(>=\) 65 & 7,732 \\ \bottomrule
\end{tabular}
\label{tbl_3}
\end{table}

In our trauma dataset, the sex feature was categorized into two groups: male (M) and female (F). In addition, the age group was classified into two categories: individuals under 65 years of age and those 65 years and older. We present the sizes of these four groups in Table \ref{tbl_3}.

\subsubsection{Employment Data}
We acquired our third dataset from Kaggle.com, titled ``Employability Classification of Over 70,000 Job Applicants" \cite{tankha_70k_job_applicants}. This dataset contains comprehensive information regarding job applicants and their employability outcomes, compiled from a variety of sources, including job portals, career fairs, and online applications. The primary objective of this dataset is to assist organizations in evaluating candidates' suitability for different roles by offering insights into the factors that influence employability. It contains both personal and professional attributes, such as age, sex, education level, years of coding experience, and previous salary. The target variable indicates whether an applicant has been hired (employed or not). With over 70,000 instances, this dataset provides a diverse sample of applicants from various industries, skill levels, and job functions.

We selected this dataset for two critical reasons. First, its large and diverse sample size allows us to test the concept of our faithfulness metrics in a corporate HR setting. Second, and more importantly, the presence of sensitive attributes like gender and age makes it a prime candidate for bias/fairness analysis. This allows us to explore how model compression might affect bias in a hiring context, where unfair predictions could lead to discriminatory outcomes and significant legal and ethical consequences for an organization. The dataset's straightforward purpose of assisting in recruitment and its potential for ``bias detection'' make it an ideal test case for our research. The dataset comprises a range of key features, including categorical variables such as `Age', `Gender', and `EdLevel', which provide essential demographic context, as well as continuous variables like `PreviousSalary' and `YearsCodePro', which offer insights into an individual's professional experience and compensation. However, this dataset may be susceptible to inherent biases, specifically in terms of gender or age inequalities, which could significantly influence the employability analysis. Such biases might affect the model’s capability to generalize and could unintentionally reinforce discriminatory hiring practices that adversely affect underrepresented groups.

\begin{table}[ht]
\caption{Bias Subgroups in the Employment Dataset}
\centering
\begin{tabular}{lr} \hline
\textbf{Bias Group} & \textbf{Size} \\ \hline 
ManOrNot = Man & 63,598 \\ 
ManOrNot = Not Man & 4,623 \\[.35em]
GenderedOrNot = Gendered & 66,959 \\
GenderedOrNot = Not Gendered & 1,262 \\[.35em]
Age \(<\) 35 & 45,859 \\ 
Age \(>\) 35 & 22,362 \\ \hline
\end{tabular}
\label{tbl_2}
\end{table}

In the dataset, the gender feature was categorized into three distinct groups: Man, Woman, and NonBinary. To analyze bias, we divided this column into two distinct categories. The first category was Man vs. Not Man (encompassing both women and non-binary individuals). The second category was Gendered (comprising men and women) versus Not Gendered (which included non-binary individuals). After preprocessing the dataset and removing instances with missing values and outliers, we arrived at a total of 68,221 instances. Table \ref{tbl_2} presents the six subgroups and their corresponding sizes.

\subsection{Model Architecture and Training Configuration}
We developed our models using Artificial Neural Networks (ANNs), implemented via the Keras API within the TensorFlow framework \cite{developers2022tensorflow}. Each model architecture consisted of multiple fully connected layers with ReLU (Rectified Linear Unit) activation functions applied to the hidden layers. The ReLU function was chosen for its computational efficiency and ability to mitigate vanishing gradient issues, thereby enabling more effective learning of complex nonlinear patterns \cite{relu_activation_builtin}. The output layer employed a sigmoid activation function \cite{wanto2017use}, which maps outputs to the (0,1) interval, making it well-suited for probabilistic interpretation in binary classification tasks.

We compiled our models using the Adam optimization algorithm, which is widely used for its fast convergence and adaptive learning rate capabilities. We set the loss function to binary\_crossentropy, which is common in binary classification problems, and assess the difference between predicted probabilities and actual binary labels.

\subsection{Data Splitting and Validation Strategy}
To assess model generalization and robustness, we used a holdout-based validation strategy. Each dataset was divided into 80\% training and 20\% testing sets. Additionally, within the training set, 10\% of the data was further set aside for validation purposes. This three-way split allowed us to monitor and tune model performance on unseen data during training while keeping a final test set for unbiased evaluation. All splits were randomized and stratified to ensure that the class distribution remained consistent.

\subsection{Baseline Model Tuning}
For each of the three datasets used in our study, we developed a separate baseline model. We performed hyperparameter tuning for each dataset to customize the model architecture and training configuration for optimal performance. This process involved frequent testing across a grid of configurations, including the number of hidden layers, the number of units per layer, the learning rate, the batch size, and the dropout rate (where applicable). Ultimately, we selected the final baseline models based on the highest average validation accuracy and consistency across different runs.

\subsection{Model Compression Techniques}
After developing our baseline models, we implemented two techniques for model compression—quantization and pruning. Our goal was to evaluate their impact on performance and fairness as well as their alignment with the baseline models using our novel metrics. For both compression techniques, we utilized the TensorFlow Model Optimization toolkit \cite{tensorflow_model_optimization} offered by TensorFlow. This comprehensive suite of tools is designed to optimize machine learning models for efficient deployment and execution. The toolkit supports various techniques aimed at reducing latency and lowering inference costs for both cloud and edge devices, such as mobile phones and IoT devices. It allows for the deployment of models on edge devices that have constraints related to processing power, memory, power consumption, network usage, and model storage space \cite{tensorflow_model_optimization}. Additionally, the toolkit enables the optimization of execution for both existing hardware and specialized accelerators.

\subsubsection{Quantization}
For quantization, we implemented the \textit{quantization-aware training (QAT)} techniques to ensure minimal loss of accuracy during the model’s optimization phase. QAT simulates the effects of quantization during training, enabling the model to adapt to reduced-precision representations. Specifically, we simulated 8-bit fixed-point quantization for both weights and activations. After training the quantization-aware model, we further fine-tuned it for a limited number of epochs to recover any accuracy loss caused by quantization noise.

\subsubsection{Pruning}
For pruning, we used the \textit{magnitude-based pruning} algorithm implemented in TensorFlow’s Model Optimization Toolkit. This approach removes weights (links) with the smallest magnitudes under the assumption that they contribute the least to the model’s performance. We applied a polynomial decay schedule to progressively increase the pruning sparsity from an initial level to a final target value. By gradually transitioning from lower to higher sparsity levels, we can effectively simplify the model while maintaining its performance. This approach not only improves efficiency but also conserves computational resources during deployment. The pruned model was also fine-tuned post-pruning to recover performance. This technique reduces the number of non-zero parameters, resulting in lower inference latency and memory usage, which is particularly beneficial for deployment on edge devices. For the COMPAS dataset, we began with an initial sparsity of 50\% and gradually increased it to a final sparsity of 80\%. In the case of the Employment dataset and the Trauma dataset, we trained our pruned model starting at an initial sparsity of 85\% and ultimately reached a final sparsity of 95\%.

%\subsection{Experimental Setup}
\subsection{Metric Assessment and Discussion Strategy}
To address our goal of justifying and demonstrating novel metrics for assessing the faithfulness of compressed models, we provide definitions and justifications for our metrics along with demonstrations of their use on the above datasets and compression techniques. We first do so with the established metrics of size and accuracy. Then we make the case for both model agreement as well as model bias as additional informative novel metrics to assess model faithfulness. 

\section{Model Size and Accuracy}\label{sec:accuracy}
In relevant literature, the most commonly used metric for evaluating compression is the model size. Meanwhile, the performance of machine learning models, including compressed ones, is primarily measured by their predictive accuracy. While this paper advocates for a more comprehensive assessment of trustworthiness, these two metrics remain essential as a baseline for measuring the efficiency and performance of compressed models.

Model size refers to the amount of memory or storage space required to store the model's parameters. For Artificial Neural Networks (ANNs), this typically depends on the number of layers, the number of neurons in each layer, and the precision of the numerical format used to represent the weights and biases (for example, 32-bit floating-point versus 8-bit integer) \cite{buciluǎ2006model}. When deploying models on edge devices with limited resources, minimizing model size often becomes a primary objective. A smaller model consumes less storage, requires less memory during inference, and can lead to faster loading times, all of which are crucial for applications on smartphones, embedded systems, and Internet of Things (IoT) hardware. The necessity of this metric is fundamentally rooted in the practical constraints of real-world deployment environments.
Model accuracy measures the proportion of correct predictions made by the model on a given dataset. It is a primary indicator of a model's predictive performance and its ability to generalize to unseen data. Although we argue that accuracy alone is insufficient for assessing the faithfulness of a compressed model, it remains a critical metric. A significant drop in accuracy after compression may render the model unusable for most applications, regardless of its size. 

\begin{figure}
    \centering
    \includegraphics[width=.7\linewidth]{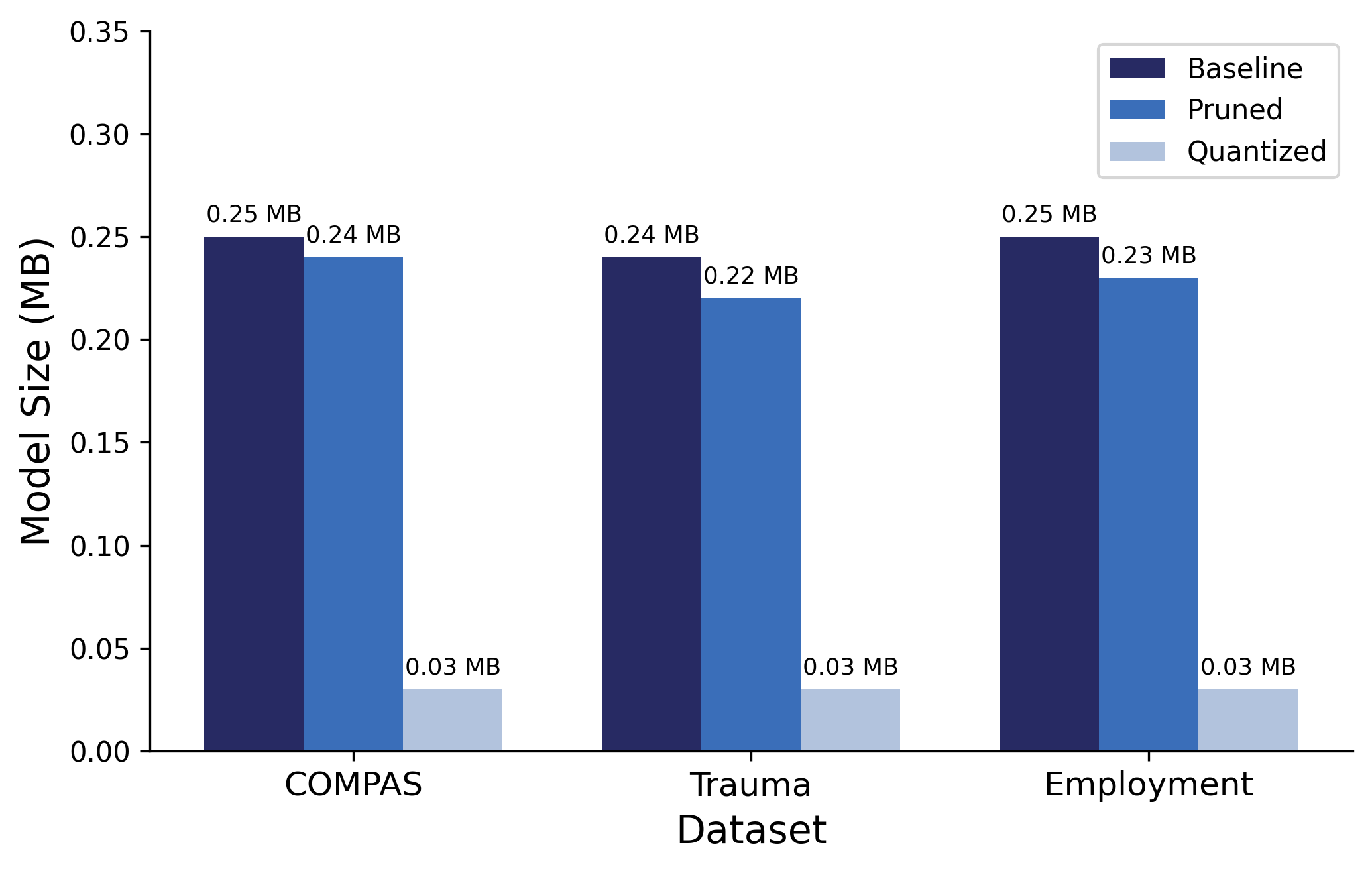}
    \caption{Difference in model size per dataset.}
    \label{fig:size_comp}
\end{figure}

\subsection{Applying Size and Accuracy Metrics}
For our research, we utilized three different datasets: the COMPAS Recidivism Racial Bias dataset, the Kentucky Trauma Triage dataset, and the Employment (HR) dataset. For each dataset, we established a baseline model using an artificial neural network (ANN) and subsequently compressed the tuned ANN model through quantization and pruning to evaluate the compressed models. For each compression technique and the baseline, we measured and recorded the resulting model size in megabytes (MB) using a single run. Figure \ref{fig:size_comp} illustrates the size differences of each compression method compared to the baseline across all datasets.

For accuracy, we will provide an example with the COMPAS dataset and discuss our observations. To ensure the reliability of our results, we conducted all experiments (other than size) ten times, collecting results for each iteration along with their aggregates.  

In each iteration, we began by splitting our dataset into training and testing sets, using an 80/20 train-test split. We then scaled our features using the StandardScaler library from Python's Scikit-Learn. Following this, we built our baseline artificial neural network (ANN) model, which was tuned to the COMPAS dataset. Our baseline model consisted of a fully connected, seven-layer ANN, with the widest layer containing 136 neurons. To reduce overfitting, we implemented dropout in our baseline model. Once the model was built, we recorded predictions and calculated metrics, including accuracy, precision, recall, and F1-score. The baseline model for the COMPAS dataset achieved an average accuracy of approximately 83\%. 

\begin{table*} [h]
\caption{Classification metrics (mean and standard deviation) for baseline, quantized, and pruned models.\\\textit{Note:} Sample standard deviation is in parentheses.}
\centering
\resizebox{\textwidth}{!}{%
\begin{tabular}{llllll}
\toprule
\textbf{Dataset} & \multicolumn{1}{c}{\textbf{Classification}} & \multicolumn{4}{c}{\textbf{Results}} \\
\cmidrule{3-6}
 & \textbf{Model} & \textbf{Accuracy} & \textbf{Precision} & \textbf{Recall} & \textbf{F1-score} \\
\midrule
\multirow{3}{*}{COMPAS} 
 & Baseline & 0.826 (0.0112) & 0.818 (0.0170) & 0.829 (0.0269) & 0.823 (0.0121) \\
 & Quantized & 0.820 (0.0077) & 0.808 (0.0205) & 0.829 (0.0231) & 0.818 (0.0081) \\
 & Pruned & 0.708 (0.0080) & 0.719 (0.0166) & 0.660 (0.0396) & 0.687 (0.0173) \\
\midrule
\multirow{3}{*}{Trauma} 
 & Baseline & 0.840 (0.0032) & 0.758 (0.0098) & 0.732 (0.0148) & 0.745 (0.0060) \\
 & Quantized & 0.838 (0.0030) & 0.760 (0.0128) & 0.724 (0.0175) & 0.741 (0.0056) \\
 & Pruned & 0.836 (0.0035) & 0.775 (0.0126) & 0.687 (0.0187) & 0.728 (0.0081) \\
\midrule
\multirow{3}{*}{Employment} 
 & Baseline & 0.778 (0.0041) & 0.771 (0.0063) & 0.821 (0.0130) & 0.795 (0.0042) \\
 & Quantized & 0.778 (0.0036) & 0.770 (0.0089) & 0.822 (0.0179) & 0.795 (0.0047) \\
 & Pruned & 0.778 (0.0038) & 0.776 (0.0034) & 0.812 (0.0052) & 0.793 (0.0034) \\
\bottomrule
\end{tabular}%
}
\label{tab:classification-results}
\end{table*}

Next, we applied quantization to the baseline model using the TensorFlow Model Optimization Toolkit, specifically the quantize\_model module in Keras and then recompiled the compressed model. Additionally, we created a pruned model by employing the magnitude pruning technique in Keras and recompiling it to enhance performance. We then recorded predictions and performance metrics (accuracy, precision, recall, and F1-score) for both compressed models. This process was repeated for ten iterations, and we calculated the average and standard deviation for each metric. Table \ref{tab:classification-results} presents the average and standard deviation of the recorded performance metrics from all ten iterations for each model across each dataset. 

\subsection{Predicting Accuracy of Compressed Models}\label{sec:predAcc}
After calculating the accuracy of our baseline and compressed models, we aimed to determine whether we could use a validation set to estimate the potential accuracy of a model. This is important for all our metrics. We need to be able to reliably anticipate metric values on future unseen data using a validation set in order for the metric to be useful for determining whether or not a compressed model is acceptable.

\begin{figure}
    \centering
    \includegraphics[width=.85\linewidth]{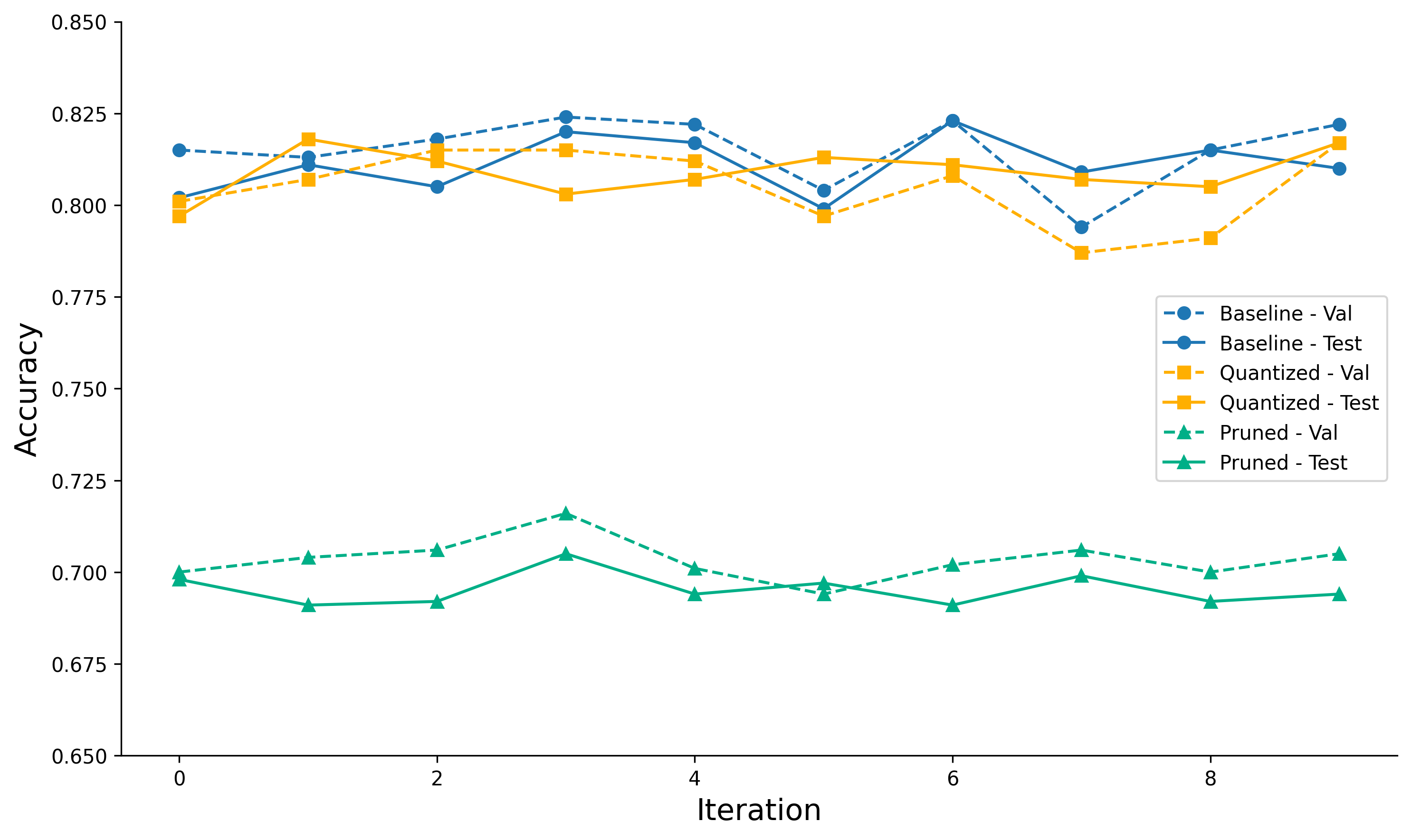}
    \caption{Change in validation and test accuracy over 10 iterations on the COMPAS dataset for Baseline, Quantized, and Pruned models.}
    \label{fig:change_acc_compas}
\end{figure}

To conduct this test, we needed a validation split of our data in addition to the training and test sets. Ultimately, we decided on a final split of 70\% for the training set, 15\% for the validation set, and 15\% for the test set. In each loop iteration, we recorded the accuracy of both the baseline and compressed models on the validation and test sets to assess the correlation between the two. Figure \ref{fig:change_acc_compas} shows the change in accuracy from validation to test set for Baseline, Quantized, and Pruned models on the COMPAS data. For each of our datasets, we also calculated the Root Mean Squared Error (RMSE) to assess the change in accuracy of each model. RMSE is a widely used metric that quantifies the average magnitude of the errors between predicted values and actual values. It places greater emphasis on larger errors because of the squaring process involved. This makes RMSE particularly useful in applications where identifying and penalizing significant deviations is crucial. Furthermore, RMSE is mathematically convenient for optimization in machine learning algorithms and is commonly utilized across various domains. We report the RMSE for each model on each dataset in Table \ref{tab:rmse-comparison-accuracy}.

\begin{table}[ht]
\caption{Root Mean Squared Error (RMSE) and Mean Absolute Percentage Error (MAPE) on Accuracy for Baseline, Quantized, and Pruned models across datasets.}
\centering
% Table a: RMSE
\begin{tabular}{lllllll}
\toprule
\textbf{Dataset} & \multicolumn{3}{c}{\textbf{RMSE}} & \multicolumn{3}{c}{\textbf{MAPE}}\\
\cmidrule(r{6pt}){2-4} \cmidrule(l{6pt}){5-7}
 & \textbf{Baseline} & \textbf{Quantized} & \textbf{Pruned} & \textbf{Baseline} & \textbf{Quantized} & \textbf{Pruned}\\
\midrule
COMPAS     & 0.0088 & 0.0108 & 0.0095 & 0.9\% & 0.9\% & 1.9\%\\
Trauma     & 0.0044 & 0.0055 & 0.0095 & 0.3\% & 0.5\% & 1.0\% \\
Employment & 0.0129 & 0.0127 & 0.0094 & 1.8\% & 1.6\% & 1.2\%\\
\bottomrule
\end{tabular}
\label{tab:rmse-comparison-accuracy}
\end{table}
To complement the insights gained from RMSE, we also calculated the Mean Absolute Percentage Error (MAPE) for our models. While RMSE provides a measure of the error's magnitude in the original units, MAPE offers a more intuitive, relative perspective by expressing the average error as a percentage \cite{kim2016new}. It is calculated by averaging the absolute percentage differences between the forecasted (in our case, validation) and actual (test) values.
This metric is particularly valuable for its straightforward interpretation; a MAPE of 5\%, for example, means the model's predictions are, on average, off by 5\%. By reporting MAPE in Table \ref{tab:rmse-comparison-accuracy}, alongside RMSE, we provide a more complete picture of model performance. This dual approach allows us to consider both the magnitude of the errors (RMSE) and their relative significance (MAPE), thus providing a more complete way to evaluate how effectively our validation set results predict the accuracy of the final test set.
\subsection{Analysis and Discussion}
The observed results show a clear connection among model size, predictive accuracy, and the selection of compression techniques. Our analysis focuses on understanding these trade-offs and their implications for model deployment.

A primary goal of model compression is to reduce size for deployment on resource-constrained devices. As shown in Figure \ref{fig:size_comp}, quantization provides a substantial benefit in this regard. Across all three datasets (COMPAS, Trauma, and Employment), the baseline models were approximately 0.24 to 0.25 MB in size. Quantization drastically reduced this size to just 0.03 MB, resulting in an approximately 88\% size reduction. On the contrary, pruning led to only a slight reduction in model size, decreasing from 0.25 MB to 0.24 MB for the COMPAS dataset, with similarly minor reductions for the other datasets. This outcome emphasizes an important distinction between model complexity and model size. While pruning creates a sparse model with lower computational complexity, it does not guarantee a smaller model size without specialized handling. A 32-bit float representing a zeroed-out weight still occupies 32 bits of storage, whereas a quantized 8-bit integer occupies only 8. Therefore, if the sole objective is minimizing storage footprint, quantization is logically superior in this context. However, if the primary goal is to reduce model complexity for reasons such as interpretability or reducing computational operations, pruning might still be considered.

\begin{figure}
    \centering
    \includegraphics[width=.7\linewidth]{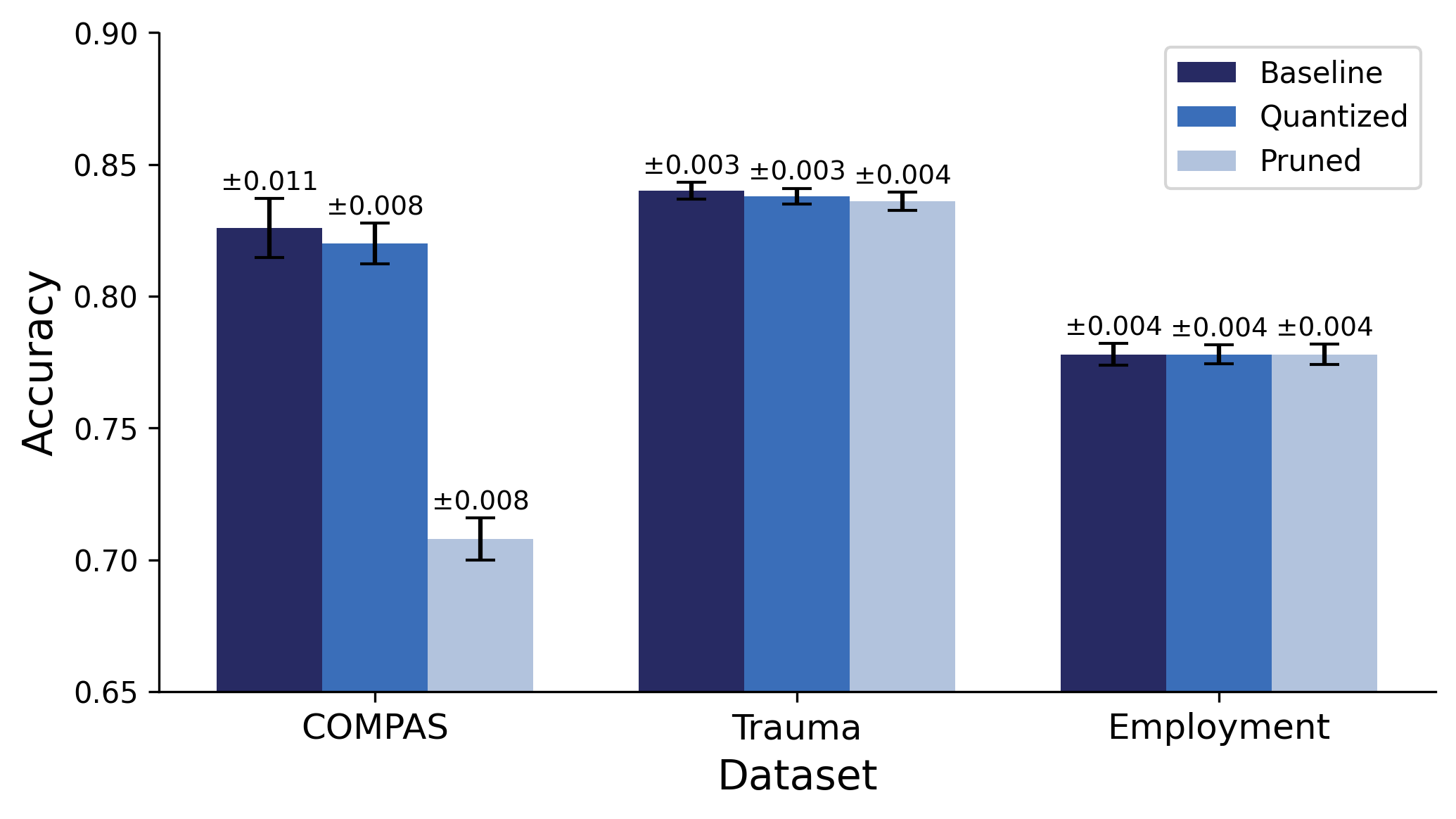}
    \caption{Average model accuracy per dataset}
    \label{fig:acc_comp_dataset}
\end{figure}

However, size reduction is only acceptable if it does not significantly compromise model performance. The data presented in Table \ref{tab:classification-results} and visualized in Figure \ref{fig:acc_comp_dataset} reveal a noticeable difference in the performance impact of the two compression methods. The quantized models consistently maintained performance nearly identical to their respective baselines. For instance, on the COMPAS dataset, the baseline accuracy was 0.826 (±0.0112), while the quantized model achieved 0.820 (±0.0077). Similar levels of stability were found in precision, recall, and F1-score across all datasets. This indicates that, for this dataset, quantization can achieve significant model compression without a noticeable loss in predictive power. In contrast, pruning resulted in a severe degradation of model performance for the COMPAS dataset. The accuracy dropped from 0.826 to 0.708, a decrease of nearly 12\%. This was accompanied by considerable drops in precision (0.818 to 0.719), recall (0.829 to 0.660), and F1-score (0.823 to 0.687). However, the performance degradation for the Trauma and Employment datasets was not as severe. These results indicate that the predictive performance of compressed models may vary depending on the type of dataset used.

Furthermore, our analysis of performance over multiple iterations provides insights into model stability and predictability. As shown in Figure \ref{fig:change_acc_compas} for the COMPAS dataset, the validation and test accuracies for both the baseline and quantized models track each other closely across all ten iterations. This suggests that for these models, accuracy on a validation set is a reliable predictor of accuracy on an unseen test set, which is a desirable characteristic for model development and deployment. The low Root Mean Squared Error (RMSE) and Mean Absolute Percentage Error (MAPE) values reported in Table \ref{tab:rmse-comparison-accuracy} for these models would further confirm this stability. 

This analysis shows that quantization is an effective method for reducing model size while maintaining accuracy, precision, and recall. On the other hand, pruning, as implemented with these datasets, provides a worse trade-off; the performance costs far exceed its minimal size advantages. This highlights an important point: choosing the proper compression technique is crucial. A careless application of a method can lead to unfavorable outcomes.

These findings, based on standard performance metrics, lay the groundwork for our primary investigation. Since compression techniques can have varying effects on accuracy, it is essential to also evaluate their impact on more direct measures of trustworthiness, such as model agreement and fairness. This ensures that a compressed model is not only efficient and accurate but also reliable and faithful to the predictive behavior of the original model.

\section{Model Agreement} \label{sec:agreement}
Understanding how compressed models compare to their uncompressed counterparts is crucial for evaluating the faithfulness and reliability of model compression. While accuracy remains a dominant evaluation metric, it can obscure meaningful discrepancies in decision-making behavior between models. Moreover, two models may achieve the same accuracy yet behave differently when applied to specific data subgroups or decision boundaries. Therefore, assessing model agreement provides a more direct way to determine the faithfulness of compression. This metric measures the extent to which the predictions of a compressed model align with those of the original model, providing insight into whether compression leads to systematic shifts or inconsistencies in predictions. Differences between the two models' predictions can highlight areas of reduced faithfulness, especially if they disproportionately impact sensitive or underrepresented data.

\subsection{Applying Agreement Metrics}
\begin{figure}
    \centering
    \includegraphics[width=.6\linewidth]{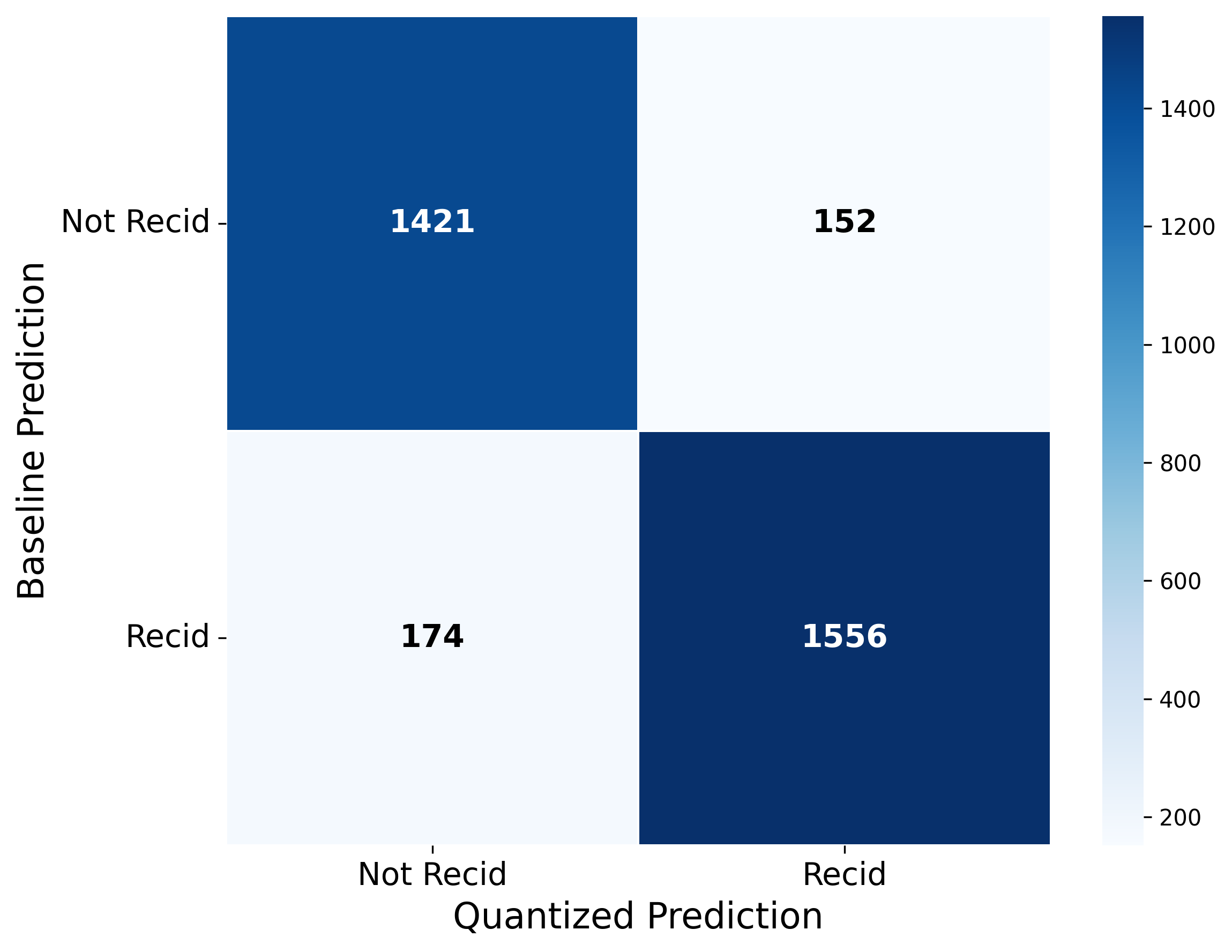}
    \caption{Agreement Statistic Matrix - COMPAS (Baseline vs Quantized model)}
    \label{fig:AgreementStat_COMPAS_quant}
\end{figure}
In this section, we quantify the level of agreement between each of the compressed models and the baseline model, treating the baseline as the reference point. For each of the ten iterations of test set predictions, we first recorded the predicted labels from the baseline model, as well as those from the quantized and pruned models. Next, we calculated the number of True Positives (TP), True Negatives (TN), False Positives (FP), and False Negatives (FN) for each compressed model relative to the baseline. This provided us with the statistics on how many instances of each class the models agreed or disagreed on. Figure \ref{fig:AgreementStat_COMPAS_quant} illustrates an example of the agreement statistic matrix for the quantization technique applied to the COMPAS dataset. The top left section of the matrix shows how many instances were agreed upon as `Not Recid,' (1421) while the bottom right section shows their agreement on `Recid' (1556). The top right and the bottom left sections of the matrix represent disagreements between the two models for that particular iteration. 

\begin{table*}[ht]
\caption{Agreement-based performance metrics (mean and standard deviation) between Quantized/Pruned models and Baseline.\\\textit{Note:} Sample standard deviation is in parentheses.}
\centering
\resizebox{\textwidth}{!}{%
\begin{tabular}{llllll}
\toprule
\textbf{Dataset} & \multicolumn{1}{c}{\textbf{Compressed}} & \multicolumn{4}{c}{\textbf{Results (Agreement with Baseline)}} \\
\cmidrule{3-6}
 & \textbf{Model} & \textbf{Accuracy} & \textbf{Precision} & \textbf{Recall} & \textbf{F1-score} \\
\midrule
\multirow{2}{*}{COMPAS} 
 & Quantized & 0.896 (0.0103) & 0.890 (0.0190) & 0.901 (0.0189) & 0.895 (0.0117) \\
 & Pruned    & 0.740 (0.0112) & 0.763 (0.0267) & 0.691 (0.0356) & 0.724 (0.0119) \\
\midrule
\multirow{2}{*}{Trauma} 
 & Quantized & 0.960 (0.0035) & 0.942 (0.0212) & 0.928 (0.0237) & 0.935 (0.0057) \\
 & Pruned    & 0.924 (0.0072) & 0.912 (0.0172) & 0.837 (0.0286) & 0.872 (0.0111) \\
\midrule
\multirow{2}{*}{Employment} 
 & Quantized & 0.970 (0.0076) & 0.973 (0.0192) & 0.974 (0.0220) & 0.973 (0.0068) \\
 & Pruned    & 0.954 (0.0053) & 0.968 (0.0102) & 0.950 (0.0149) & 0.959 (0.0042) \\
\bottomrule
\end{tabular}%
}
\label{tab:agreement-with-baseline}
\end{table*}

Next, we calculated the average \textit{Agreement Accuracy}, \textit{Agreement Precision}, \textit{Agreement Recall}, and \textit{Agreement F1-score}, along with their standard deviations based on the values from the agreement statistic matrix. We also measured the rates of agreement and disagreement across different demographic subgroups to determine whether the compressed models maintained consistent decision-making across various populations (discussed in Section \ref{sec:bias}). We report the Mean and Standard Deviation for the agreement accuracy, precision, recall, and F1-score from all ten iterations in Table \ref{tab:agreement-with-baseline}.

\subsection{Testing for Statistical Significance}
To assess whether the observed changes in agreement behavior due to compression are statistically significant, we used the chi-squared test and calculated p-values. To determine the p-value for a chi-squared test, one first formulates null and alternative hypotheses regarding the association between categorical variables \cite{pandis2015calculating}. The core of the test involves calculating the chi-squared \( (\chi^2) \) statistic, which quantifies the discrepancy between observed \textit{(O)} and expected \textit{(E)} frequencies across all categories using the formula \cite{utah_chi_square} \[ \chi^2 = \sum \frac{(O - E)^2}{E}\] Besides this, the degrees of freedom \textit{(df)} are also determined based on the dimensions of the data table. The p-value is then derived from the chi-squared distribution using the calculated \( \chi^2 \) statistic and its corresponding \textit{df} \cite{utah_chi_square}. Finally, this p-value is compared to a chosen significance level (e.g., \(\alpha=0.05\)) to decide whether to reject or accept the null hypothesis.

\begin{figure}
    \centering
    \includegraphics[width=.75\linewidth]{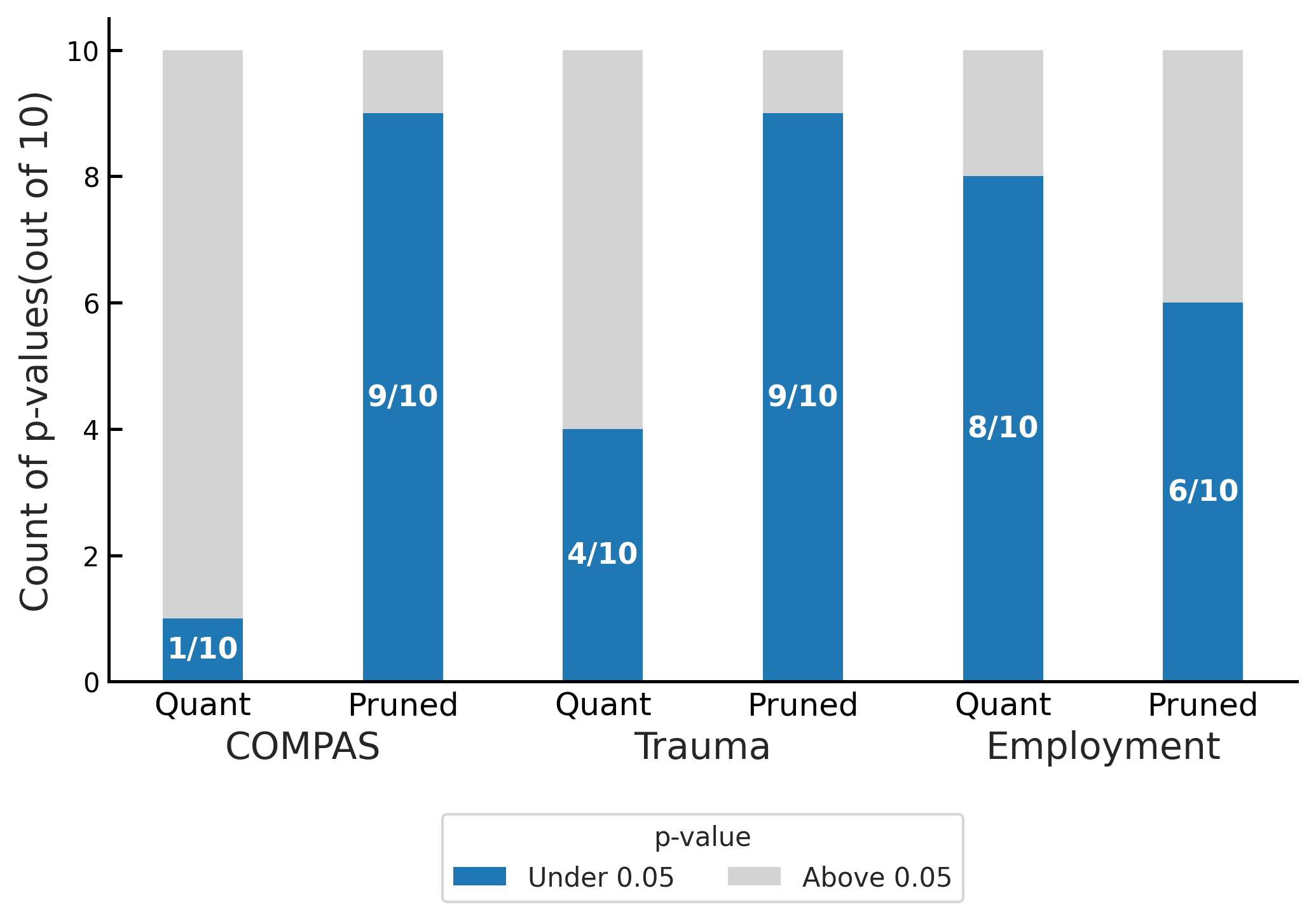}
    \caption{Number of times p-values were above/below threshold for each compressed model across all datasets \textit{Note: Points below threshold (0.05) are considered `bad' compressions}}
    \label{fig:p-val_Agreement}
\end{figure}

For our experiment, we used the built-in chi2\_contingency \cite{scipy2024chi2} function provided in the scipy.stats module in Python to calculate the p-values. For that, we first had to construct a contingency table from our agreement statistics to provide as input to the function. A contingency table involves two or more categories, showing how different groups relate to each other by counting the occurrences of each combination. One category is positioned across the top (columns) and the other down the side (rows), with each cell indicating the frequency of specific pairs co-occurring. Totals at the end of each row and column reveal how often each category appears. We built our contingency table by aggregating the different values in the agreement statistics confusion matrix. The resulting table had the class on one side and the models (Baseline and Compressed) on the other side. Once we had our contingency table, we ran it through the chi2\_contingency function. This returned the p-value for the agreement statistic corresponding to the compression method used in that particular loop. If we found the p-value to be statistically significant (i.e.,\ $\leq 0.05$), we rejected the null hypothesis (that there was no relationship between the compression and changes in agreement) and concluded that the compression is not faithful, or in other words, `bad.' We also performed this analysis ten times to assess how many compressions had agreement statistics that were statistically significantly not faithful compared to the baseline. Figure \ref{fig:p-val_Agreement} shows the number of times p-values recorded in the ten runs were above/below the threshold for each compressed model based on their agreement with the baseline. 

\subsection{Predicting Model Agreement}
\begin{figure}
    \centering
    \includegraphics[width=.85\linewidth]{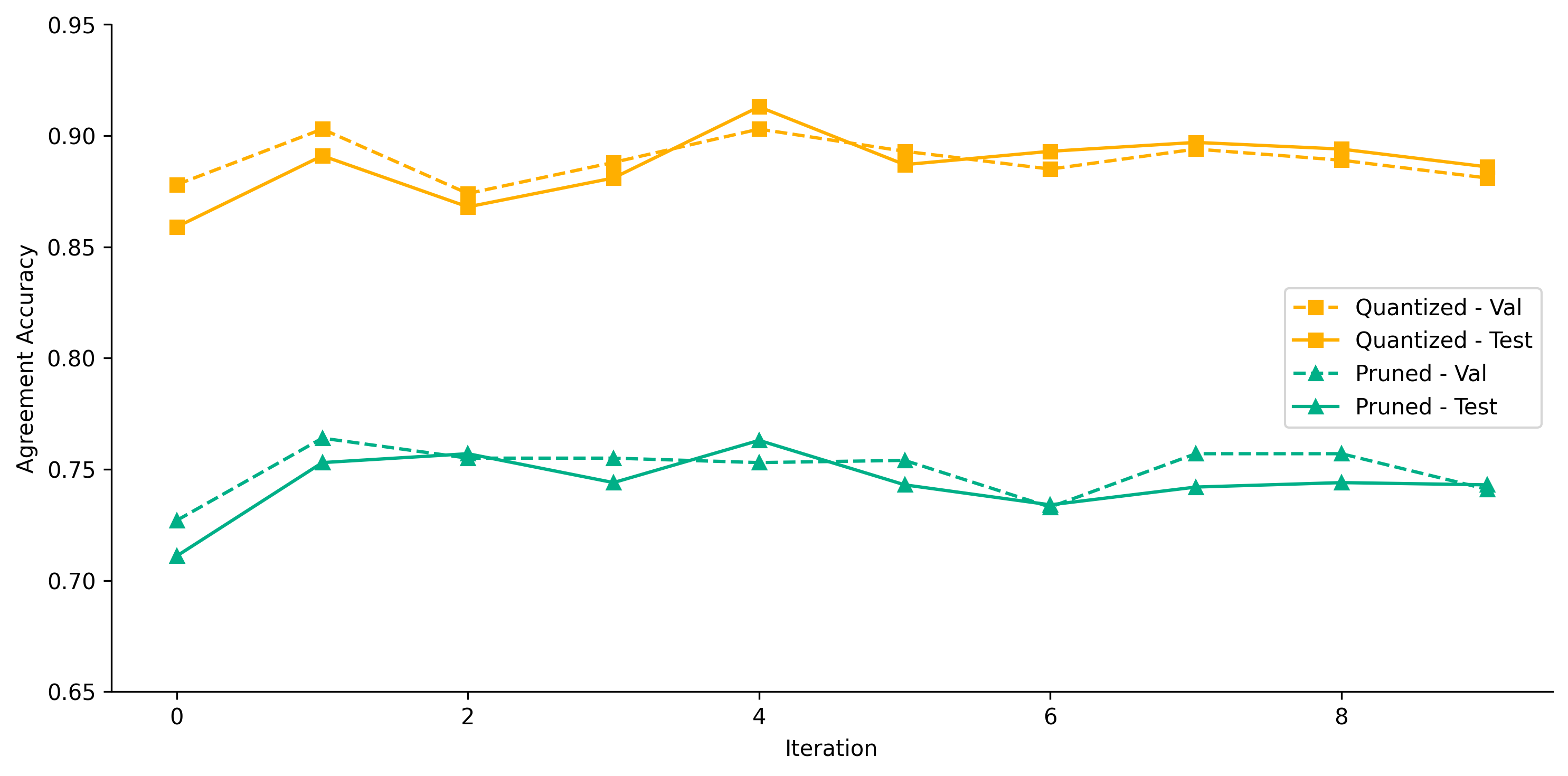}
    \caption{Change in validation and test \textit{agreement accuracy} over 10 iterations on the COMPAS dataset for Quantized, and Pruned models.}
    \label{fig:change-agreement-compas}
\end{figure}

\begin{table}[ht]
\caption{Root Mean Squared Error (RMSE) and Mean Absolute Percentage Error (MAPE) on Agreement Accuracy for Baseline, Quantized, and Pruned models across datasets.}
\centering
\begin{tabular}{lllll}
\toprule
\textbf{Dataset} & \multicolumn{2}{c}{\textbf{RMSE}} & \multicolumn{2}{c}{\textbf{MAPE}}\\
\cmidrule(r{6pt}){2-3} \cmidrule(l{6pt}){4-5}
 & \textbf{Quantized} & \textbf{Pruned} & \textbf{Quantized} & \textbf{Pruned}\\
\midrule
COMPAS     & 0.0092 & 0.0001 & 0.9\% & 1.2\% \\
Trauma     & 0.0039 & 0.0063 & 0.3\% & 0.7\% \\
Employment & 0.0010 & 0.0030 & 0.1\% & 0.3\% \\
\bottomrule
\end{tabular}
\label{tab:rmse-mape-comparison-agreement}
\end{table}

Similar to our approach for model accuracy, we aimed to assess whether we could predict the ``goodness" or faithfulness of the compressed model, as well as its agreement with its uncompressed counterpart, using validation sets. To conduct this test, we used a split of 70\% for training, 15\% for validation, and 15\% for testing. In each loop, we constructed our baseline and compressed models. We then recorded the agreement accuracy for both the baseline and the compressed models on the validation and test sets. We report our observations on the change in agreement accuracies for the validation and test sets of Baseline, Quantized, and Pruned models on the COMPAS dataset in Figure \ref{fig:change-agreement-compas}. For each of our datasets, we also report the change in agreement accuracy in Table \ref{tab:rmse-mape-comparison-agreement}, quantifying the error using both RMSE and Mean Absolute Percentage Error (MAPE).  

\begin{figure}
    \centering
    \includegraphics[width=.7\linewidth]{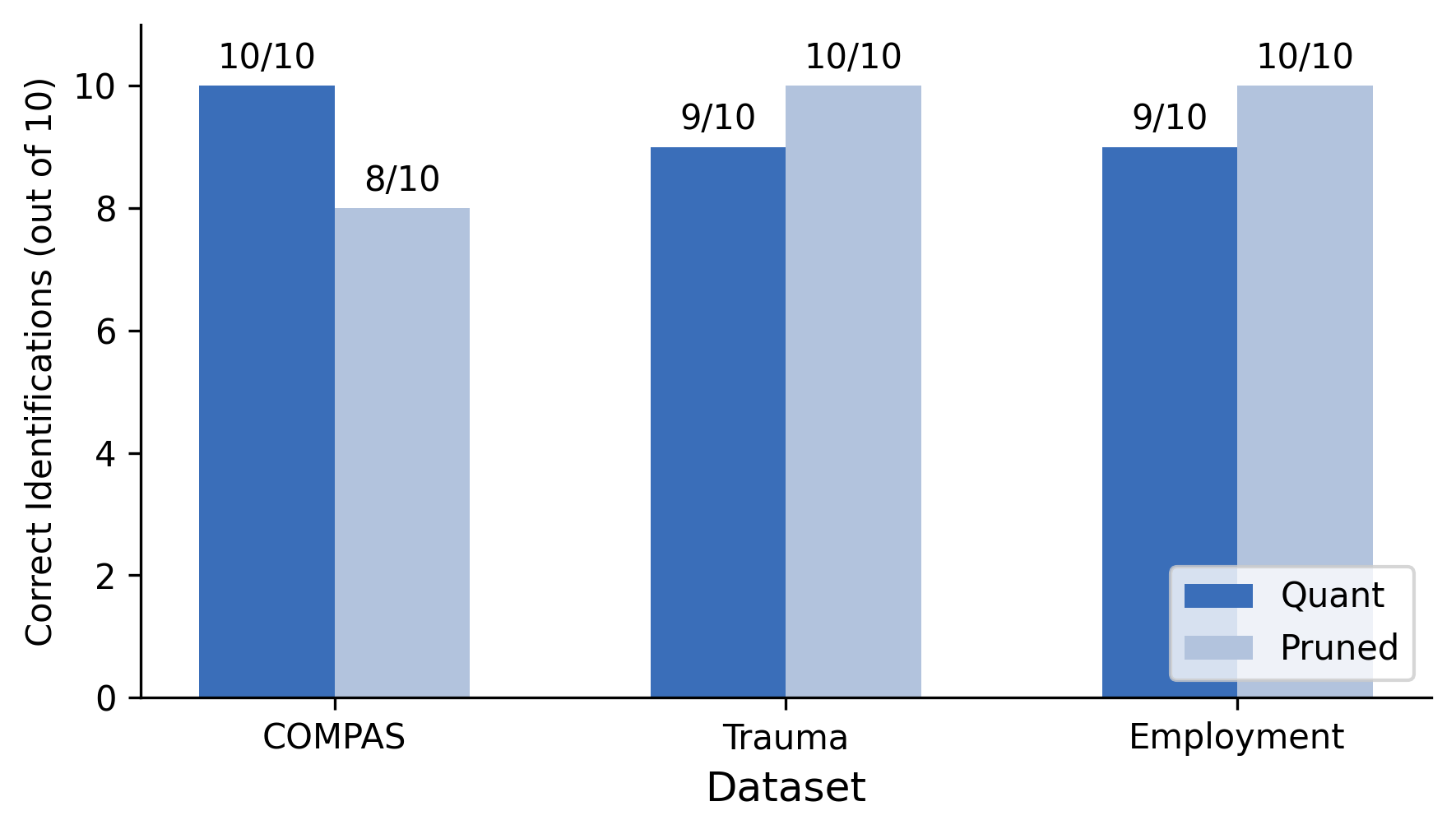}
    \caption{Validation set accuracy for predicting Model Agreement}
    \label{fig:val_set_p-val}
\end{figure}

Next, we applied the same methods outlined in Section \ref{sec:predAcc} to record the p-values for both compression techniques, using the validation set first and then the test set. Finally, we compared the statistical significance of the results from the validation set with those from the test set to determine how often we could predict whether the compressed model would remain faithful to the baseline model. Figure \ref{fig:val_set_p-val} shows the number of times the validation set accurately identified the type of compression (bad or not) out of ten runs for each compressed model across all datasets.

\subsection{Analysis and Discussion}
The analysis of model agreement reveals a more direct measurement of compression faithfulness than accuracy metrics alone can provide. By treating the baseline model’s predictions as the ground truth, we can directly measure how much a compressed model’s decision-making behavior has changed.
The agreement metrics in Table \ref{tab:agreement-with-baseline} mostly mirror the performance trends seen in the accuracy analysis. For the COMPAS dataset, the quantized model maintains a high agreement accuracy of 0.896, indicating that its predictions align with the baseline’s in nearly 90\% of instances. In contrast, the pruned model’s agreement accuracy drops to 0.740, confirming that its poor predictive performance is a result of it fundamentally disagreeing with the baseline model’s decisions. For the Trauma and Employment datasets, both compression techniques achieve very high agreement scores (ranging from 0.924 to 0.970), suggesting the compression process had a less disruptive effect on the final predictions for these models.

While high agreement accuracy is essential, it doesn’t provide a complete picture. The chi-squared test adds a crucial layer of analysis by assessing whether the observed disagreements are random instabilities or indicative of a systematic, statistically significant shift in the model’s predictive distribution. The results, shown in Figure \ref{fig:p-val_Agreement}, are particularly insightful. On the COMPAS dataset, 9 out of the 10 quantized model runs produced p-values above the 0.05 significance threshold, suggesting that, in most cases, the disagreements were not statistically significant. The pruned model, however, produced statistically significant changes more frequently (in 9 out of 10 runs), aligning with its poor agreement score. The most interesting finding comes from the Employment dataset. Here, despite very high agreement accuracies for both quantized (0.970) and pruned (0.954) models, the chi-squared test overwhelmingly flagged the changes as statistically significant. For the quantized model, 8 out of 10 runs yielded a p-value below 0.05, and for the pruned model, 6 out of 10 did. This demonstrates a critical concept: a compressed model can agree with the baseline on the vast majority of cases but still introduce a non-random, systematic change in how it handles the few instances where it disagrees. Such a shift could have serious implications in practice, as it might affect a specific, small subgroup of the data, a detail that high overall agreement would hide.

Finally, we analyzed the predictability of our agreement metrics using a validation set. The predictability of agreement accuracy itself was evaluated to determine if this metric is stable between validation and testing. As illustrated for the COMPAS dataset in Figure \ref{fig:change-agreement-compas}, the agreement accuracy on the validation set closely tracks the performance on the test set across all ten iterations for both the quantized and pruned models. This visual evidence is quantitatively confirmed in Table \ref{tab:rmse-mape-comparison-agreement}. The low Root Mean Squared Error (RMSE) and Mean Absolute Percentage Error (MAPE) values across all three datasets demonstrate that the error between validation and test set agreement is minimal. This high degree of stability indicates that agreement accuracy is a reliable metric, allowing developers to confidently estimate the final agreement performance using a validation set early in the development cycle. Building on this, the experiment to predict the faithfulness via statistical significance, shown in Figure \ref{fig:val_set_p-val}, demonstrates the practical utility of the chi-squared approach. The ability to correctly predict whether the changes on the test set would be statistically significant was high across all models and datasets, ranging from 80\% to 100\% accuracy. This indicates that the chi-squared test is a stable metric that can be used reliably during the development cycle to flag potentially unfaithful compressions before final deployment.

\section{Model Bias} \label{sec:bias}
Ensuring fairness and mitigating bias are critical aspects of trustworthy machine learning, particularly when models are deployed in sensitive domains. Model compression can shift how bias manifests within a system \cite{kamal2024beyond, hooker2019compressed}. Therefore, the problem with model compression is that, while it aims to improve efficiency, the process may unintentionally alter existing biases in the original model or even introduce new ones. Compressed models can perform well on aggregate metrics but behave more unfairly towards certain groups. Therefore, beyond accuracy and agreement, it is critical also to assess whether model compression alters a model’s bias, i.e., its differential treatment of subgroups based on attributes like sex, age, race, etc. Traditional fairness metrics (e.g., Equalized Odds, Demographic Parity) can indicate whether a model’s decisions are biased, but these alone are insufficient for understanding the details of compression-induced changes. Evaluating how compression alters the bias profile of a model, beyond just measuring post-compression bias, is crucial for determining the faithfulness of compression in trust-critical applications.

\subsection{Applying Change in Bias Metrics} \label{sec:empirical-examples-bias}
\begin{figure}
    \centering
    \includegraphics[width=0.70\linewidth]{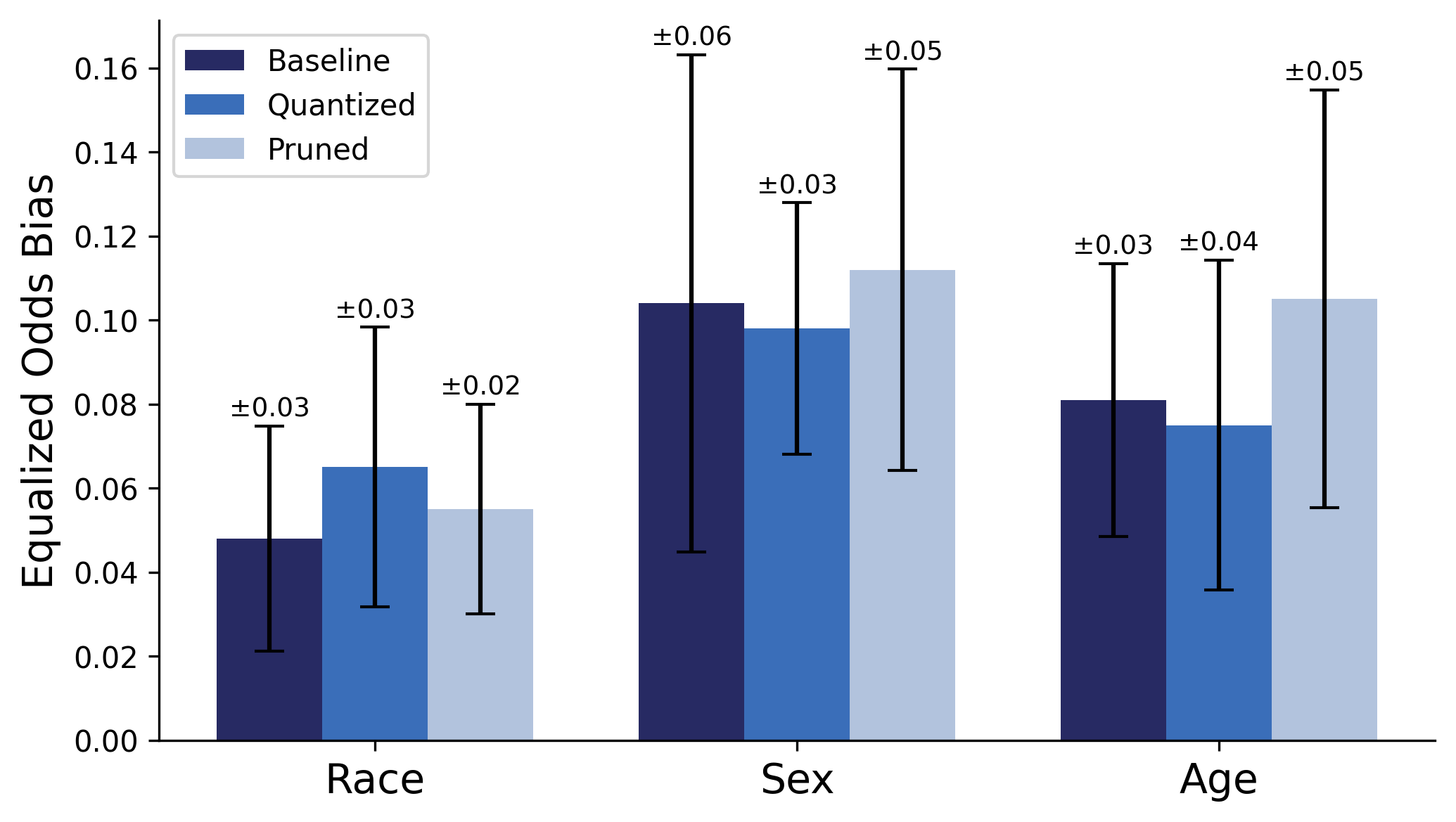}
    \caption{Average Bias for COMPAS Dataset for each Model by Demographic Subgroups}
    \label{fig:BiasCOMPAS}
\end{figure}

\begin{table*}[ht]
\centering
\caption{Bias by Equalized Odds Ratio Across Datasets, Demographics, and Models. \\\textit{Note:} Sample standard deviation is in parentheses.}
\label{tab:bias_data}
\resizebox{\textwidth}{!}{
\begin{tabular}{lllll}
\toprule
\textbf{Dataset} & \textbf{Demographic} & \multicolumn{3}{c}{\textbf{Bias by Equalized Odds Ratio}} \\ \cmidrule{3-5} 
 &  & \textbf{Baseline} & \textbf{Quantized} & \textbf{Pruned} \\ 
\midrule
\multirow{3}{*}{COMPAS} & Race & 0.048 (0.0268) & 0.065 (0.0333) & 0.055 (0.0249) \\ 
 & Sex & 0.104 (0.0591) & 0.098 (0.0299) & 0.112 (0.0477) \\
 & Age & 0.081 (0.0325) & 0.075 (0.0392) & 0.105 (0.0497) \\ 
\midrule
\multirow{2}{*}{Trauma} & Sex & 0.025 (0.0134) & 0.028 (0.0120) & 0.039 (0.0124) \\ 
 & Age & 0.081 (0.0280) & 0.077 (0.0226) & 0.081 (0.0183) \\ 
\midrule
\multirow{3}{*}{Employment} & ManOrNot & 0.026 (0.0113) & 0.032 (0.0268) & 0.044 (0.0157) \\
 & GenderedOrNot & 0.095 (0.0452) & 0.070 (0.0491) & 0.075 (0.0321) \\
 & Age & 0.043 (0.0174) & 0.042 (0.0269) & 0.018 (0.0079) \\
\bottomrule
\end{tabular}
}
\end{table*}
For this section, we evaluated the propagation of bias from the baseline model to the compressed models. We evaluated group fairness using the \textit{equalized odds} criterion, which requires that all members of the demographic subgroups have equal sensitivity (True Positive Rate) and equal specificity (True Negative Rate). When applying equalized odds, we evaluate bias within a demographic; thus, the closer the value is to zero (where zero indicates no bias), the lower the bias present. For each model and dataset, we first separated our data based on the members of each demographic group (e.g., male and female). We then calculated the sensitivity and specificity for each demographic subgroup and derived bias metrics based on their deviations. Next, we recorded the average biases and their standard deviations for each demographic subgroup. This experiment enabled us to quantify the impact of model compression on model fairness. Figure \ref{fig:BiasCOMPAS} shows the average bias as measured by Equalized Odds along with their standard deviations for the COMPAS dataset across each demographic subgroup. We also report the average bias with their standard deviations for all demographic subgroups across all datasets in Table \ref{tab:bias_data}.

However, testing the level of bias in a compressed model does not guarantee an accurate assessment of the model's faithfulness regarding compression. Similar to accuracy, equalized odds may also overlook crucial aspects of the model's faithfulness. A compressed model could achieve a similar overall bias score while fundamentally altering its decision-making for specific subgroups. Therefore, to investigate faithfulness more directly, we extended our analysis to examine bias from the perspective of model agreement.

For our agreement calculation in Section \ref{sec:agreement}, we looked at the predictions from both the baseline model and the compressed model across all test data. Similarly, to determine agreement for each demographic subgroup, we need to compare the predictions from both models for each demographic subgroup. In the three datasets we are using, most demographics are divided into two subgroups. However, some categories have more than two subgroups. For instance, in the COMPAS dataset, `Age' is divided into three subgroups: $Age < 25$, $Age = 25-45$, and $Age > 45$.  To simplify calculations, we converted all subgroups into binary categories. For COMPAS, we used $Age < 25$ and $Age \geq 25$ (the other two combined) as our two age subgroups. However, our proposed metric can be applied to more than two groups if needed.

\subsection{Testing for Statistical Significance}
\begin{table*}[ht]
\small
\caption{Contingency Tables for (a) Female, (b) Male, (c) Combined}
\centering
\begin{tabular}{cc}
% Table a: Female
\begin{tabular}[t]{lll}
\toprule
\textbf{} & \textbf{Baseline} & \textbf{Quantized} \\
\midrule
\textbf{Not Recid} & 367 & 392 \\
\textbf{Recid}     & 238 & 213 \\
\bottomrule \\
\textbf{(a) Female} \\ \\
\end{tabular}
&
% Table b: Male
\begin{tabular}[t]{lll}
\toprule
\textbf{} & \textbf{Baseline} & \textbf{Quantized} \\
\midrule
\textbf{Not Recid }& 1272 & 1307 \\
\textbf{Recid}     & 1426 & 1391 \\
\bottomrule \\
\textbf{(b) Male} \\ \\
\end{tabular}
\end{tabular}

% Table c: Combined
\begin{tabular}[t]{lll}
\toprule
\textbf{} & \textbf{Baseline} & \textbf{Quantized} \\
\midrule
\textbf{Male: Not Recid}   & 367 & 392 \\
\textbf{Male: Recid}       & 238 & 213 \\
\textbf{Female: Not Recid} & 1272 & 1307 \\
\textbf{Female: Recid }    & 1426 & 1391 \\
\bottomrule \\
\textbf{(c) Combined} \\
\end{tabular}
\label{tab:sex-male-combined}
\end{table*}

\begin{figure}
    \centering
    \includegraphics[width=0.7\linewidth]{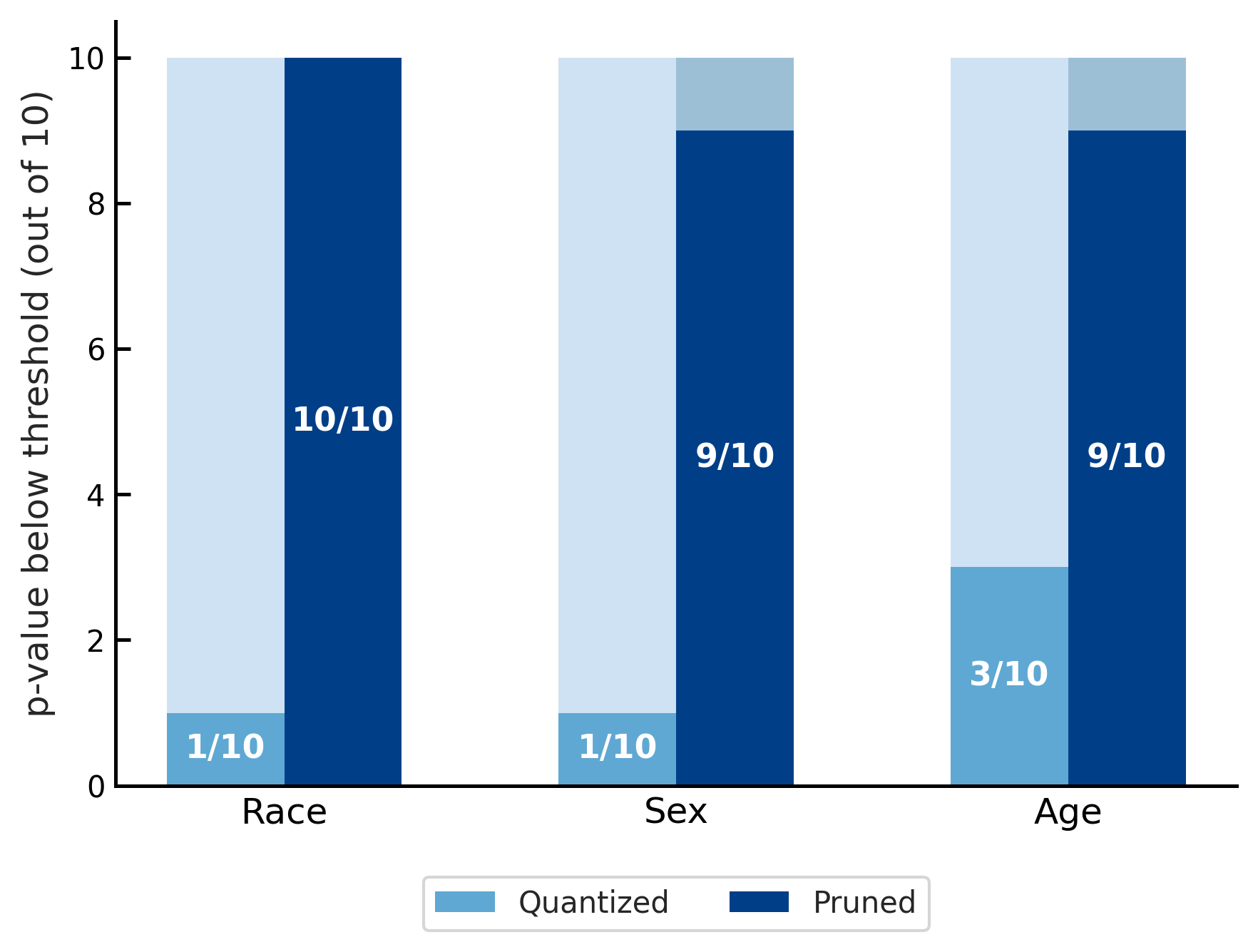}
    \caption{Number of p-values below the threshold per demographic (COMPAS Dataset)\textit{Note: Points below threshold (0.05) are considered `bad' compressions}}
    \label{fig:COMPAS_p-val_Demo_total}
\end{figure}
Next, we adapted the agreement framework to capture change in bias to create a more sensitive measure of fairness and faithfulness. To achieve this, we used the chi-squared test to determine whether a compression technique systematically alters the pattern of agreement between the baseline and the compressed model across different demographic subgroups in a statistically significant way. Since bias involves comparisons between subgroups, we started by creating separate contingency tables for each demographic subgroup and combining them into a contingency table with more dimensions. For instance, in the COMPAS dataset, for a given demographic like `Sex', we first isolated the females in the test set. We then obtained the baseline and compressed model predictions specifically for these females. Assuming the baseline predictions to be correct, we constructed a confusion matrix to determine the true positives (TP), true negatives (TN), false positives (FP), and false negatives (FN). From this, we created a 2$\times$2 contingency table solely for females (illustrated in Table \ref{tab:sex-male-combined} (a)). We repeated the procedure for males, resulting in another 2$\times$2 contingency table (Table \ref{tab:sex-male-combined} (b)). Finally, we combined both matrices to form a 2$\times$4 contingency table that included data for both males and females (Table \ref{tab:sex-male-combined} (c)). This allowed us to simultaneously assess agreement across all subgroups in a demographic feature. Using this combined table as input for the chi2\_contingency function, we calculated the p-value representing the statistical significance of any change in agreement patterns across that entire demographic. We repeated this whole process across all demographic groups and datasets for both compression methods. Figure \ref{fig:COMPAS_p-val_Demo_total} shows the number of p-values obtained from the combined contingency tables that were below the threshold (i.e.,\ $\leq 0.05$) for the COMPAS dataset.
\begin{figure}[ht]
    \centering
    \subfigure[Race - Subgroups and Total]{\label{fig:sub1}\includegraphics[width=0.45\textwidth]{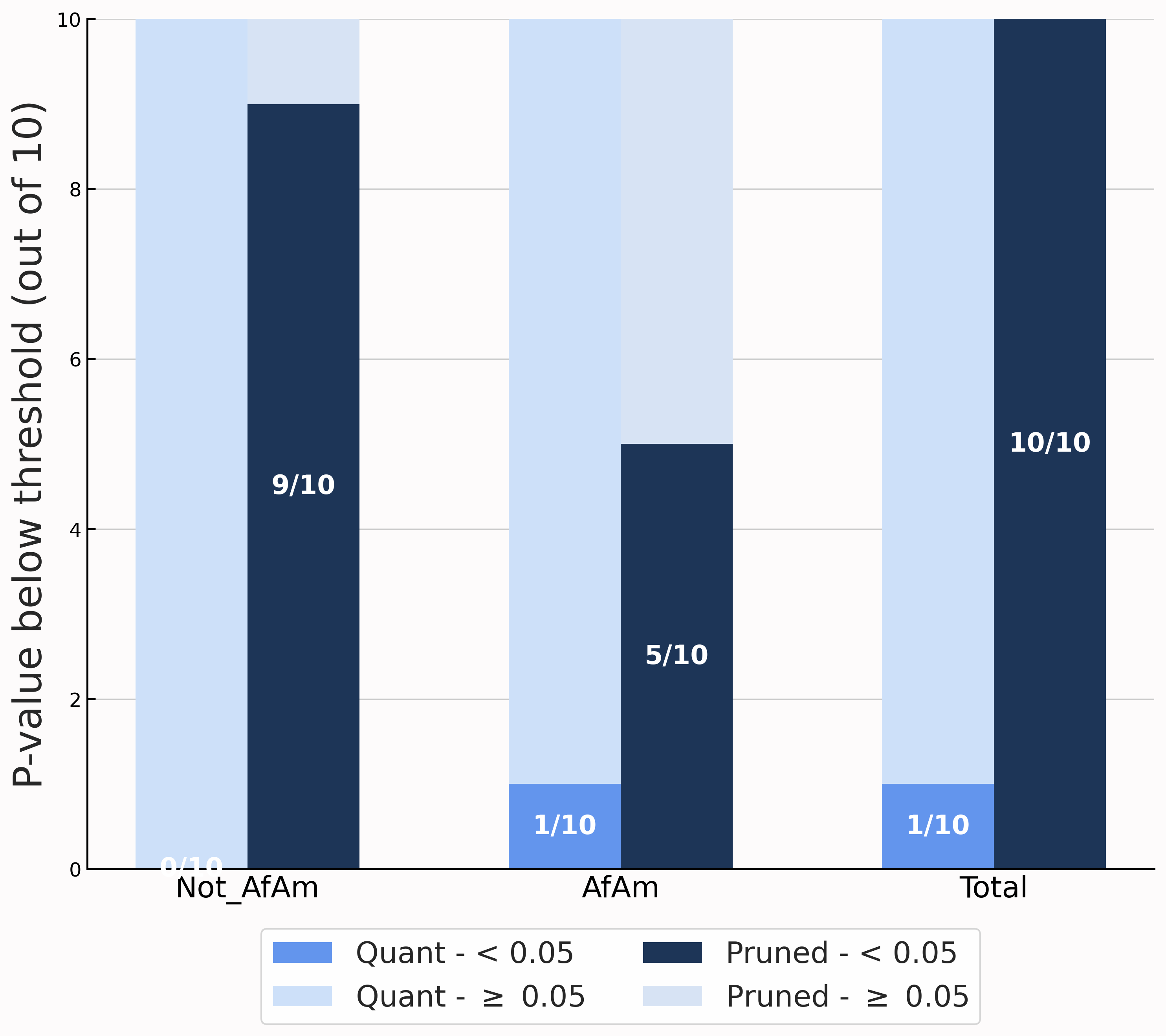}}
    \hspace{0.05\textwidth} 
    \subfigure[Sex - Subgroups and Total]{\label{fig:sub2}\includegraphics[width=0.45\textwidth]{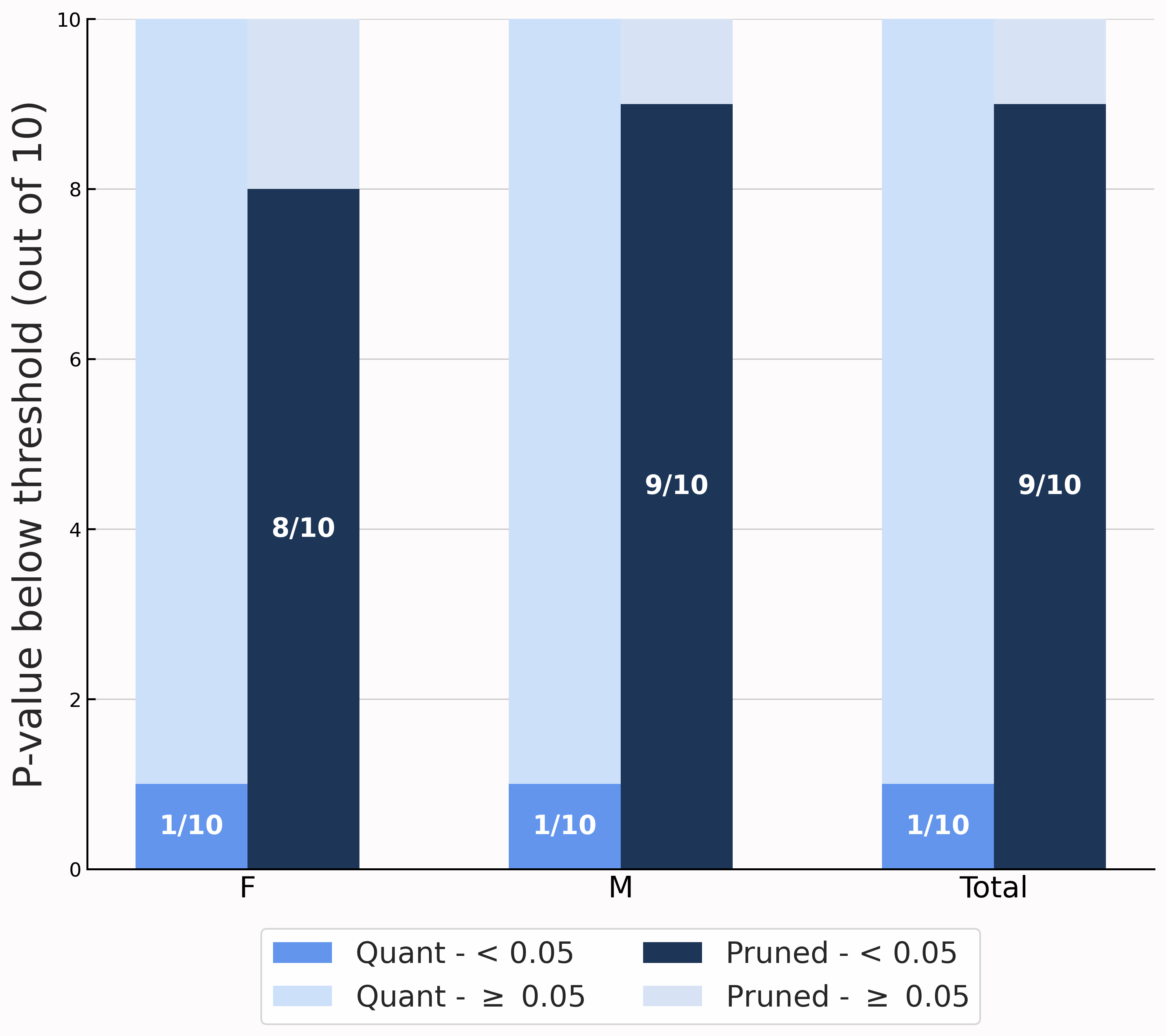}}
    \vspace{0.5em}
    \subfigure[Age - Subgroups and Total]{\label{fig:sub3}\includegraphics[width=0.45\textwidth]{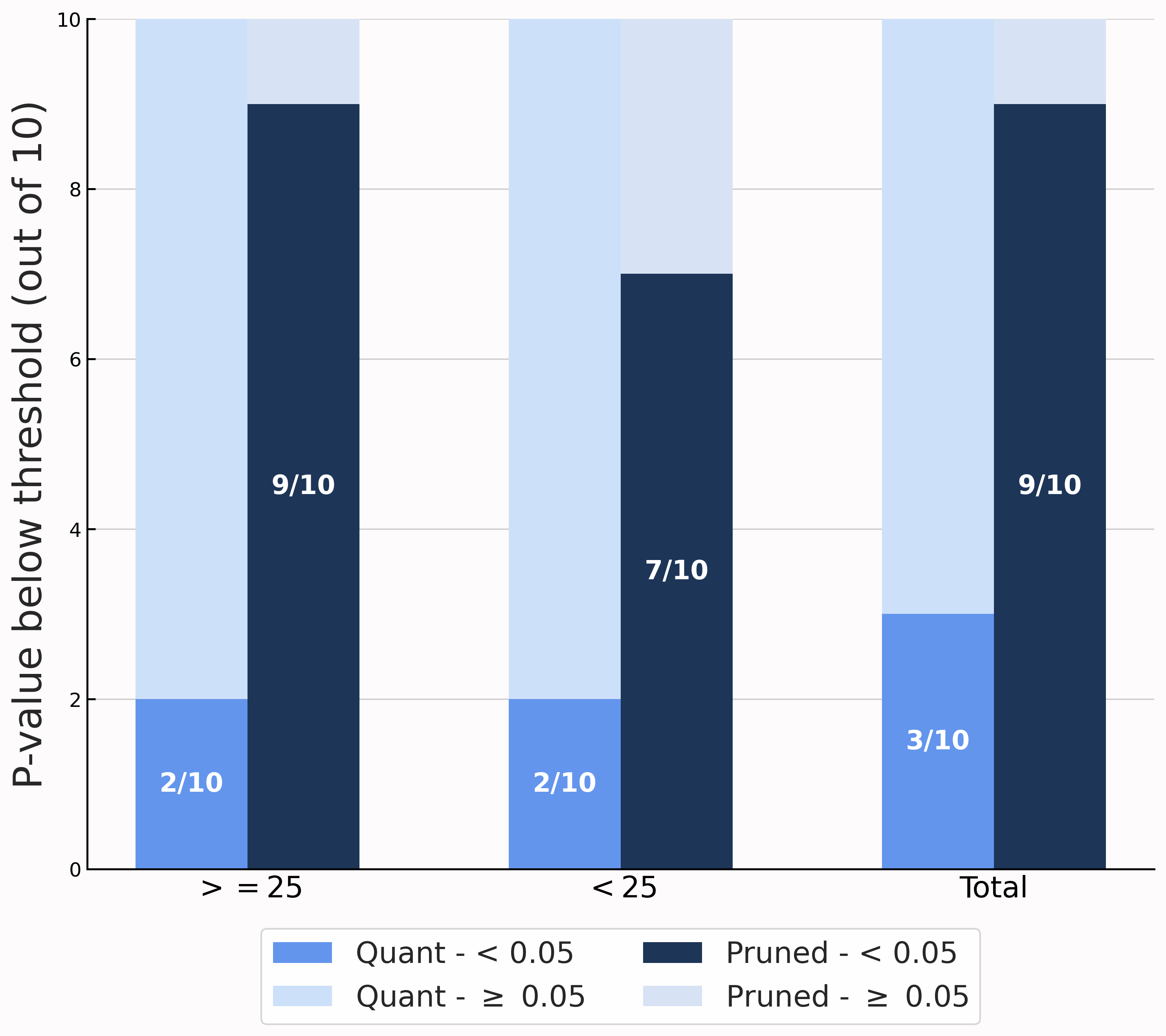}}
    \caption{Number of p-values below the threshold per demographic (COMPAS Dataset)\textit{Note: Points below threshold (0.05) are considered `bad' compressions}}
    \label{fig:COMPAS_p-val_Demo}
\end{figure}
A potential limitation of this combined method is that the results of one subgroup might obscure the results of another. To address this problem, we propose considering the subgroups separately as well as in combination. Therefore, for our analysis, we calculated p-values for demographic subgroups (such as male and female) both individually and collectively as illustrated in Figure \ref{fig:COMPAS_p-val_Demo}.

\subsection{Predicting Model Bias}
Following our analyses of accuracy and agreement, we sought to determine if a validation set could effectively predict the bias characteristics of a compressed model on the test set. For this investigation, we maintained our consistent experimental framework, utilizing a 70\% training, 15\% validation, and 15\% test data split across ten separate runs. 

\begin{figure}[ht]
    \centering
    \subfigure[Baseline]{\label{fig:bias_sub1}\includegraphics[width=0.45\textwidth]{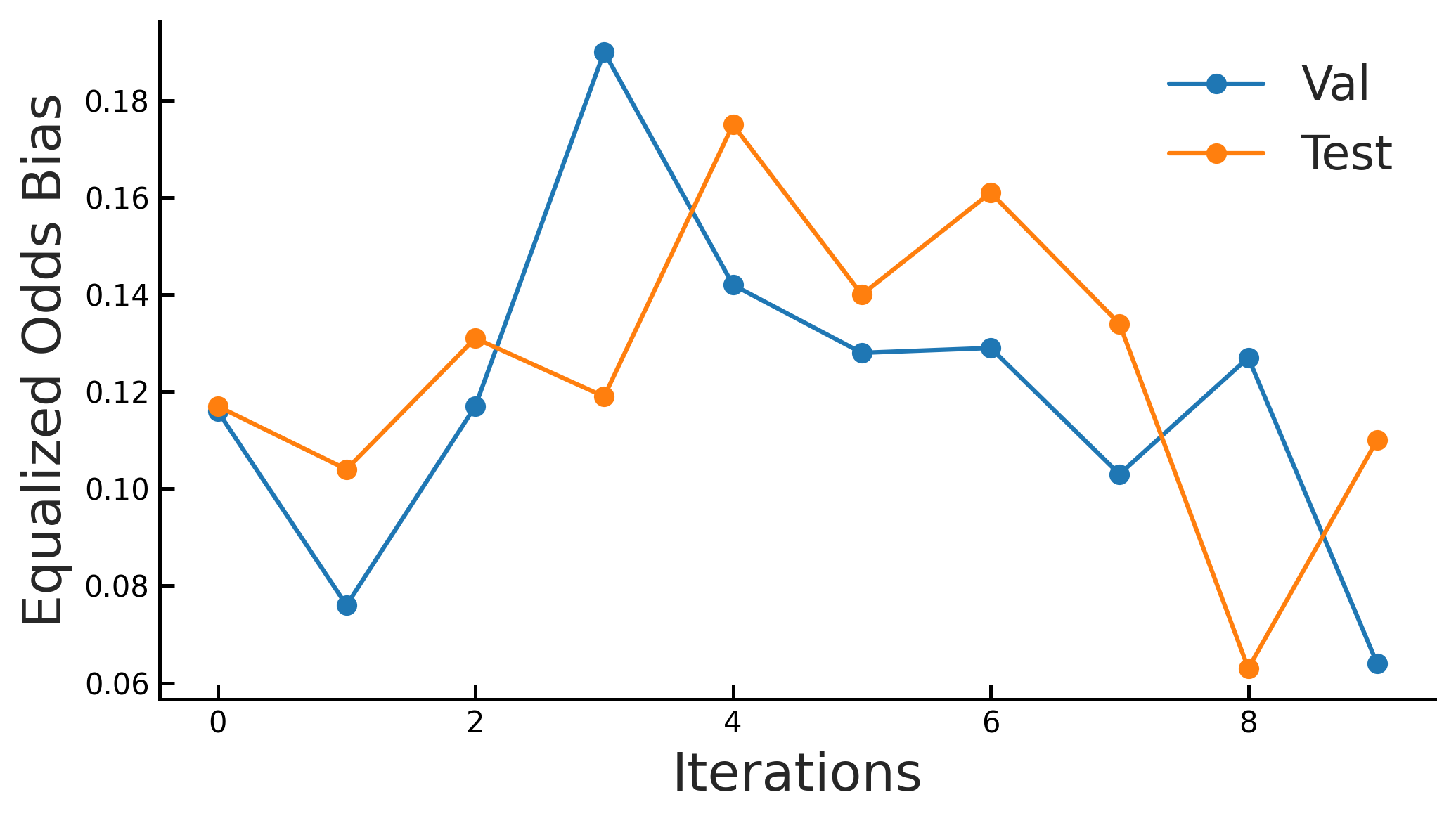}}
    \hspace{0.05\textwidth} 
    \subfigure[Quantized]{\label{fig:bias_sub2}\includegraphics[width=0.45\textwidth]{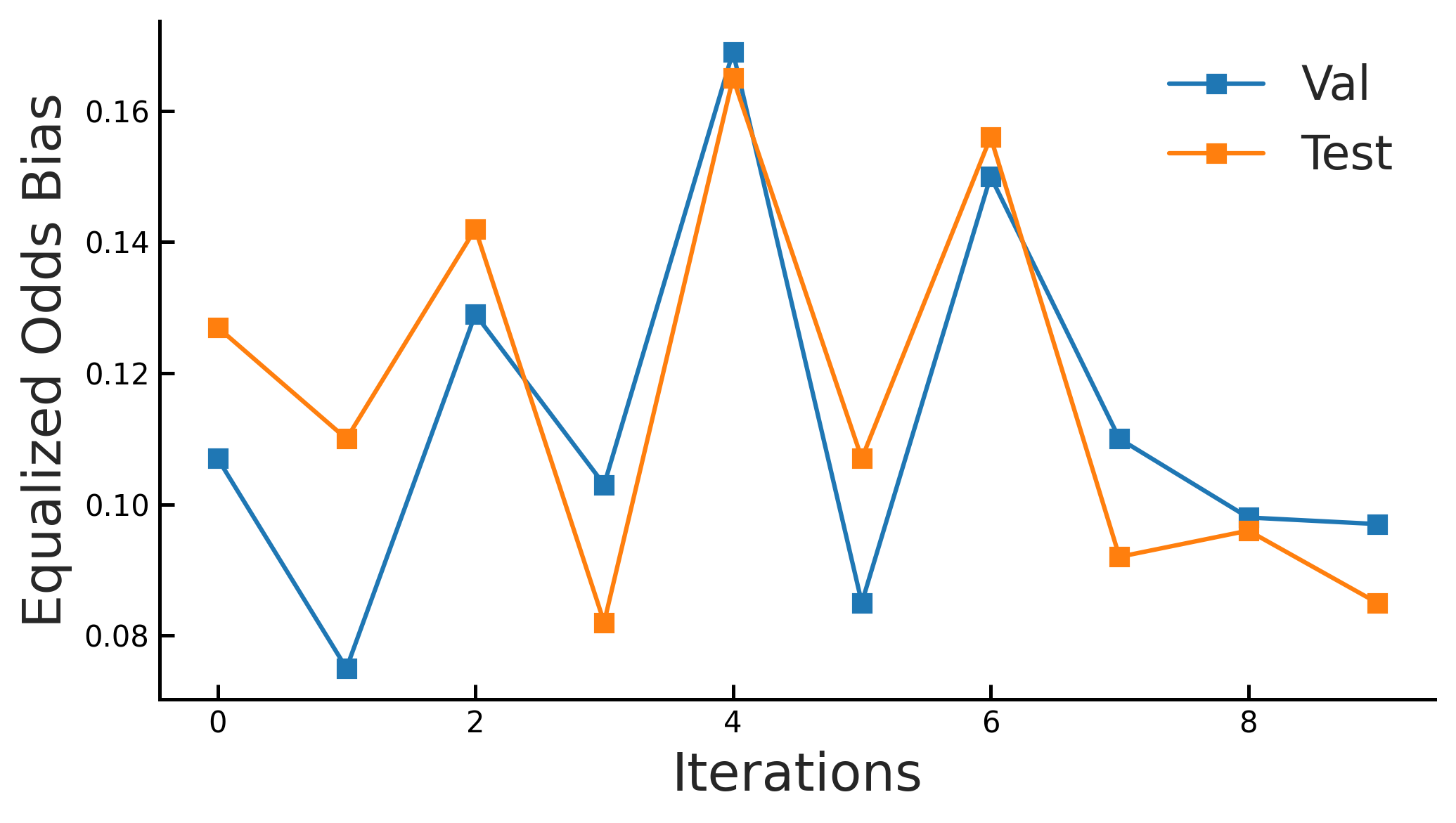}}
    \vspace{0.5em}
    \subfigure[Pruned]{\label{fig:bias_sub3}\includegraphics[width=0.45\textwidth]{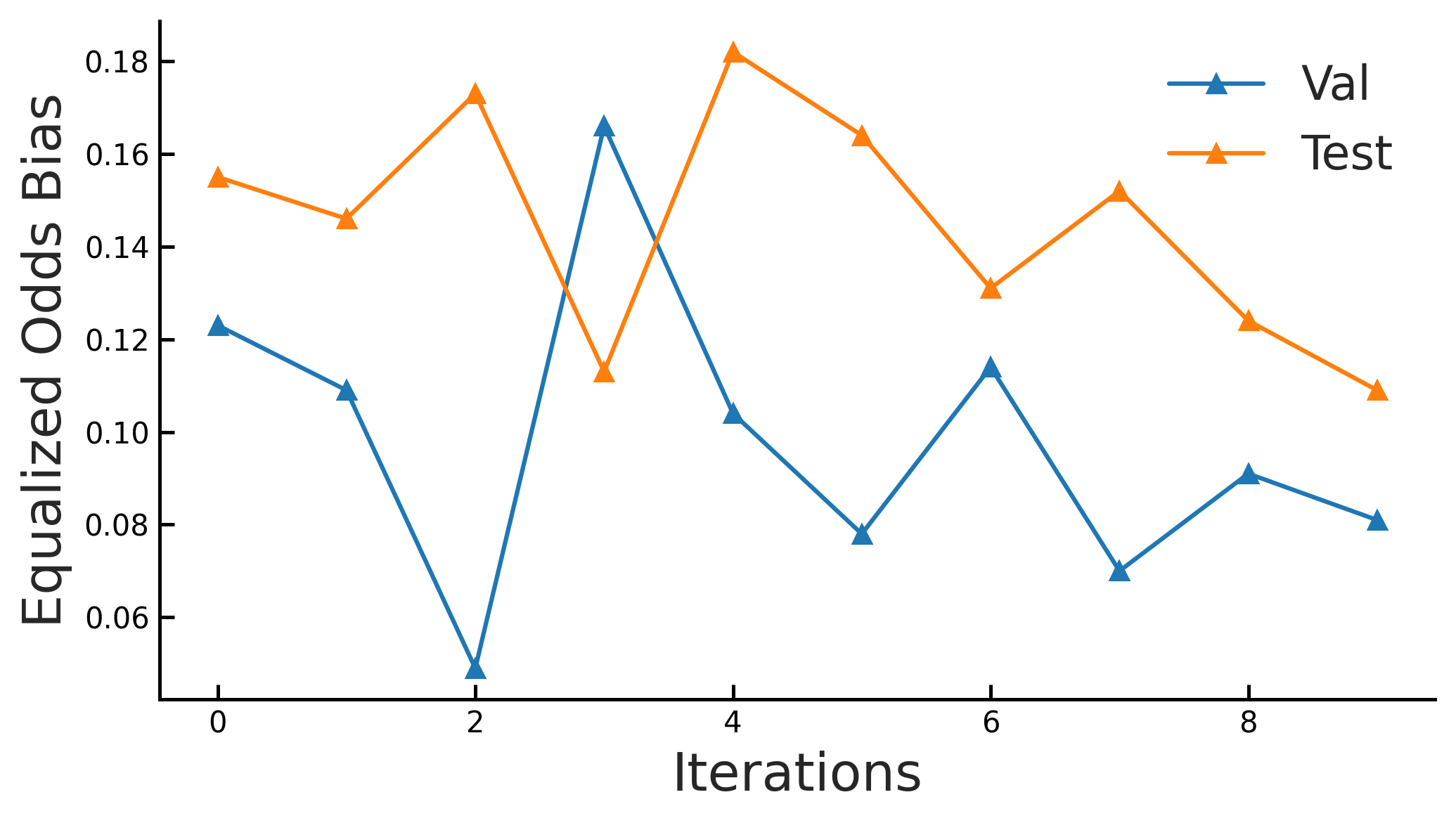}}
    \caption{Change in bias for COMPAS dataset by age subgroup for validation and test sets}
    \label{fig:bias-compas-age-valTest}
\end{figure}

\begin{table}[ht]
\caption{Root Mean Squared Error (RMSE) and Mean Absolute Percentage Error (MAPE) on change in Bias (COMPAS dataset).}
\centering
\begin{tabular}{lllllll}
\toprule
\textbf{Demographic} & \multicolumn{3}{c}{\textbf{RMSE}} & \multicolumn{3}{c}{\textbf{MAPE}}\\
\cmidrule(r{6pt}){2-4} \cmidrule(l{6pt}){5-7}
 & \textbf{Baseline} & \textbf{Quantized} & \textbf{Pruned} & \textbf{Baseline} & \textbf{Quantized} & \textbf{Pruned}\\
\midrule
Race     & 0.0282 & 0.0313 & 0.0221 & 38.4\% & 57.8\% & 28.4\% \\
Sex     & 0.0354 & 0.0507 & 0.0215 & 52.2\% & 70.0\% & 20.1\% \\
Age     & 0.0393 & 0.0180 & 0.0631 & 31.2\% & 14.5\% & 37.9\% \\
\bottomrule
\end{tabular}
\label{tab:rmse-comparison-bias}
\end{table}

Our first step was to track the change in bias from the validation to the test set using the equalized odds metric. In each of the ten iterations, we calculated the bias for all demographic groups within our datasets, first on the validation set predictions and later on the test set predictions. This was performed for the baseline, quantized, and pruned models across all datasets. By quantifying the difference in the equalized odds measurements between the two datasets, we can evaluate the stability of the bias metric itself and estimate how much it might be expected to change between validation and final testing. To visualize this change, we plotted the equalized odds bias for the validation and test sets across the ten iterations. We used separate line charts for the Baseline, Quantized, and Pruned models to ensure clarity. Figure \ref{fig:bias-compas-age-valTest} shows an example of these plots for the `Age' demographic in the COMPAS dataset, illustrating the fluctuations between validation and test set bias in each run. We then quantified the error between the validation and test set bias values using RMSE and MAPE, with the results for the COMPAS dataset reported in Table \ref{tab:rmse-comparison-bias}. However, it is essential to note a key limitation of MAPE: it can produce extremely high or misleading percentages when the actual values (the test set bias, in this case) are very close to zero because it puts a heavier penalty on positive errors than on negative errors \cite{hyndman2006another, kim2016new}. Since equalized odds values are often small, the reported MAPE values may not be a reliable measure of predictive error in this context.
\begin{figure}
    \centering
    \includegraphics[width=.7\linewidth]{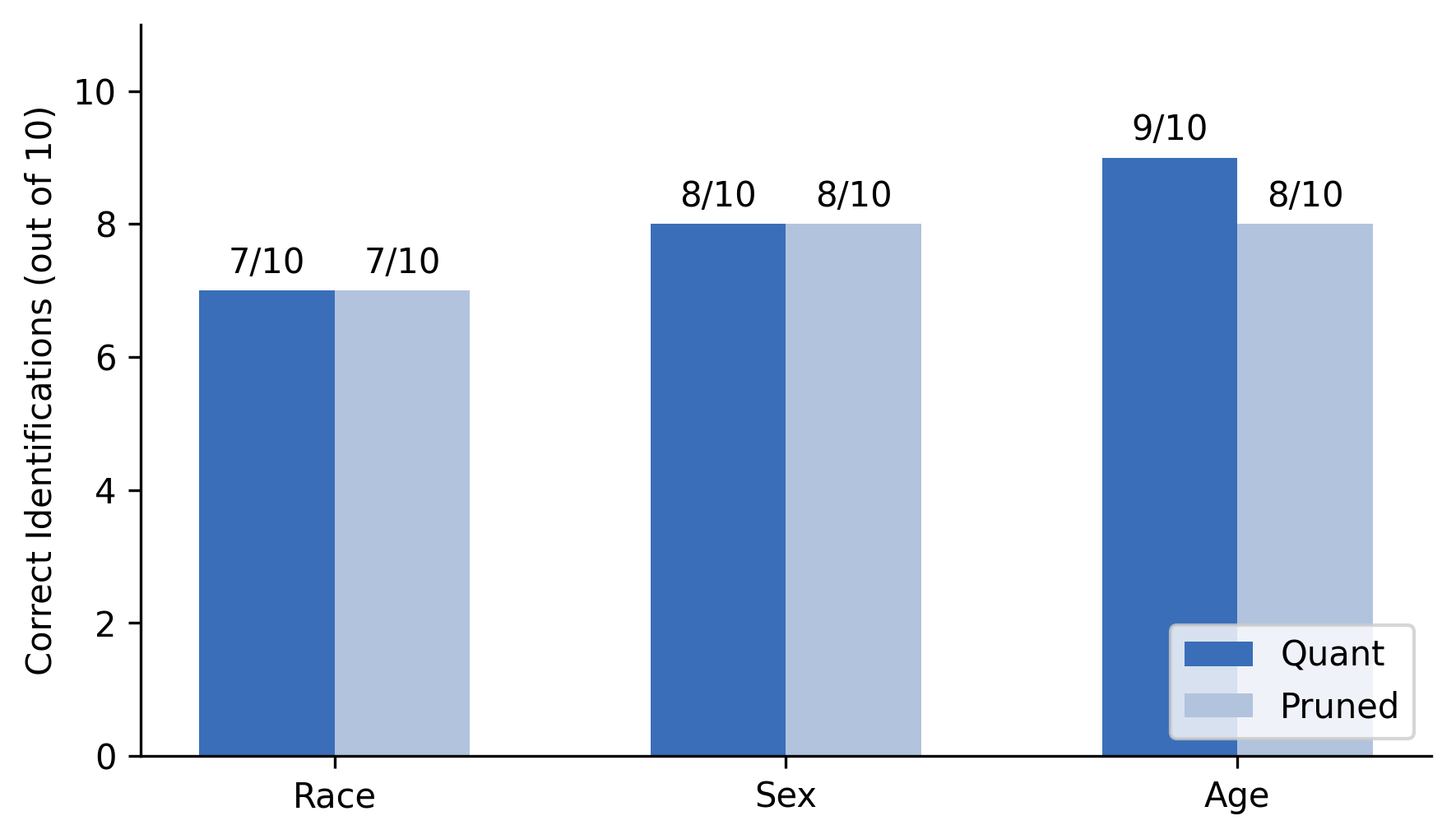}
    \caption{Validation set accuracy for predicting compression quality for each demographic subgroup (COMPAS dataset)}
    \label{fig:pval-val-test-demo-compas}
\end{figure}

\begin{table}[ht]
\centering
\caption{Validation set accuracy for the Trauma and Employment datasets}
\label{tab:correct_identifications}
\begin{tabular}{llll}
\toprule
\textbf{Datasets} & \textbf{Demographic} & \multicolumn{2}{c}{\textbf{Correct Identifications}} \\ \cmidrule{3-4} 
 &  & \textbf{Quantized} & \textbf{Pruned} \\ \midrule
\multirow{2}{*}{Trauma} & Sex & 9/10 & 10/10 \\
 & Age & 9/10 & 9/10 \\ \midrule
\multirow{3}{*}{Employment} & ManOrNot & 10/10 & 7/10 \\ 
 & GenderedOrNot & 10/10 & 8/10 \\ 
 & Age & 10/10 & 7/10 \\ \bottomrule
\end{tabular}
\end{table}
Next, to evaluate predictability from a statistical perspective, we compared the results of the chi-squared tests between the validation and test sets. Using the method of combined contingency tables for each demographic group as explained in Section \ref{sec:empirical-examples-bias}, we calculated p-values to determine if a compression technique caused a statistically significant shift in agreement patterns across subgroups. We performed this analysis first on the validation set and then on the test set for every run. By comparing the p-values from both sets, we can determine how well the validation set predicts the final statistical conclusion regarding bias and faithfulness on the test data. This allows us to see how often a flag for potential bias raised during validation holds up in the final evaluation. Figure \ref{fig:pval-val-test-demo-compas} presents a bar chart that displays the frequency with which we correctly identified the faithfulness/agreement level of each compressed model based on demographic factors for the COMPAS dataset.  We provide similar findings for the other two datasets, Trauma and Employment, in Table \ref{tab:correct_identifications}.

\subsection{Analysis and Discussion}
Our analysis aims to demonstrate that while standard fairness metrics provide a helpful baseline, a deeper statistical approach is necessary to fully assess the faithfulness of a compressed model.
First, the analysis of the Equalized Odds bias, as shown in Figure \ref{fig:BiasCOMPAS} and detailed in Table \ref{tab:bias_data}, indicates that quantization appears to be more faithful than pruning in preserving the original model's fairness profile. Across all datasets and demographic groups, the average bias of the quantized models remains very close to that of their respective baselines (e.g., for the COMPAS `Sex' demographic, the baseline bias is 0.104 and the quantized bias is 0.098). In contrast, pruning often alters the bias, sometimes increasing it (e.g., from 0.081 to 0.105 for COMPAS `Age') and exhibiting larger standard deviations, which suggests greater instability. While this indicates that quantization is the superior method for these datasets, it only describes the final state of bias, not the nature of the change that occurs.

The chi-squared test for bias agreement provides a much more sensitive measure of faithfulness. The results for the combined demographic groups, shown in Figure \ref{fig:COMPAS_p-val_Demo_total}, are clear for the pruned model. Across Race, Sex, and Age, the pruned model exhibited a statistically significant change in its agreement patterns in 9 to 10 times (out of 10) of the experimental runs. This provides strong evidence that the model systematically and unfaithfully alters its behavior. The quantized model, while far more faithful for this dataset, is not perfect. It successfully passed the test in 9 out of 10 runs for the Race and Sex demographics, but failed in 3 out of 10 runs for the Age demographic. This important finding suggests that even an effective compression technique, such as quantization, can result in subtle yet statistically significant shifts in fairness regarding specific attributes.

Furthermore, our analysis of the p-values for individual subgroups supports our dual-analysis approach, as illustrated in Figures \ref{fig:COMPAS_p-val_Demo}. For the `Race' demographic under quantization, the combined test showed a statistically significant change in only 1 out of 10 runs. Our analysis reveals that the `AfAm' subgroup entirely drove this change, since the `Not\_AfAm' subgroup showed no significant change in any of the runs (0 out of 10). This demonstrates that a combined test can obscure significant, group-specific impacts, highlighting the need for a more detailed analysis to ensure fairness for all populations.

Finally, our investigation into predicting bias highlights the challenges of forecasting fairness metrics. The line charts in Figure \ref{fig:bias-compas-age-valTest} suggest that the validation and test set biases track each other more closely for the baseline and quantized models than for the pruned model. However, the error metrics in Table \ref{tab:rmse-comparison-bias} present a more complex picture. Counterintuitively, the pruned model sometimes has the lowest RMSE, indicating that its validation-to-test error can be small even if its absolute bias is unstable. This suggests that relying solely on simple error metrics is insufficient for assessing the predictability of a model's fairness. In contrast, predicting the outcome of our statistical test proves to be more reliable. Figure \ref{fig:pval-val-test-demo-compas} shows that using a validation set to predict whether the test set would yield a statistically significant p-value was successful 7 to 9 times (out of 10) of the time. This sets our chi-squared test as a practical and reasonably reliable diagnostic tool for developers to identify potential fairness issues during the model compression workflow.

\section{Limitations}\label{sec:limitations}
This paper serves as an introduction to novel metrics to assess the faithfulness of a compressed model. While it does motivate the need for such faithfulness metrics, it does not fully examine the real-world implications of model faithlessness. Additionally, it focuses on the faithfulness of only compressed models, while the metrics could be applied to the faithfulness of any two related models, such as comparing a model built in a federated environment to one built in a more centralized manner or different models built using the same training set.

Furthermore, while this paper focused on prediction agreement and change in bias, there are additional properties that could be used to measure model faithfulness, including consistency of explanations and change in uncertainty quantification. 

\section{Conclusion}\label{sec:conclusion}
The growing use of machine learning models on resource-constrained edge devices has made model compression an essential field of study. However, the conventional focus on optimizing model size and accuracy has often overlooked critical aspects of trustworthiness, such as the faithfulness and fairness of compressed models. This paper aims to address this gap by introducing and validating a set of metrics for evaluating model compression that go beyond surface-level metrics. Our empirical investigation across multiple datasets and two standard compression techniques, quantization and pruning, yielded several key insights.

This study highlighted the limitations of relying solely on aggregate performance metrics. We introduced a statistical approach, centered on the chi-squared test, to analyze model agreement and change in bias. Our findings revealed that high post-compression accuracy does not guarantee that a model's predictive behavior remains faithful to the original. The chi-squared test successfully identified statistically significant shifts in decision-making that raw agreement scores missed. Similarly, when applied to fairness, this statistical framework proved to be a more sensitive detector of bias changes than the standard equalized odds metric. It provided definitive evidence that for our datasets, pruning systematically and unfaithfully altered the model's treatment of demographic subgroups, while also revealing subtle, group-specific fairness shifts introduced by quantization that could otherwise have gone unnoticed.

The primary contribution of this work is to provide a practical and robust methodology for assessing the faithfulness of model compression. By analyzing agreement patterns from a statistical perspective, both for the overall model and within specific demographic groups, we provide professionals with a clearer understanding of the true effects of compression. Furthermore, our results show that these statistical tests can be reasonably predicted using validation sets. This makes them a useful diagnostic tool that can be integrated into the development lifecycle to identify potentially `unfaithful' models before they are deployed.
\subsection{Future Work}
Future work will expand this framework to include additional compression techniques, such as knowledge distillation and low-rank factorization as well as additional faithfulness properties such as consistency of explanations.

Additional work will seek to explain the model differences rather than just quantify them. Also, we will evaluate whether making the threshold for validation set statistical significance more sensitive than the test set threshold can better prevent acceptance of unfaithful compressions.

\bmhead{Acknowledgements}
We would like to thank Dr. Steve Talbert at J.W. Ruby Memorial Hospital (Morgantown, WV) for providing access to the Trauma data. We also gratefully acknowledge the support from the Machine Intelligence and Data Science (MInDS) Center as well as the Department of Computer Science at Tennessee Tech University during this research.

\section*{Declarations}

\begin{itemize}
\item \textbf{Funding:} No funding was received for conducting this research.
\item \textbf{Ethics approval and consent to participate:} Not applicable.
\item \textbf{Data availability:} The COMPAS dataset and the Employment dataset are both collected from Kaggle.com. The Trauma data was obtained from the University of Kentucky. \\COMPAS - https://www.kaggle.com/datasets/danofer/compass \\Employment - https://www.kaggle.com/datasets/ayushtankha/70k-job-applicants-data-human-resource 
\end{itemize}

\bibliography{sn-bibliography}% common bib file

\end{document}